\documentclass{article} 
\usepackage{iclr2026_conference,times}

\usepackage{amsmath,amsfonts,bm}

\def\eqref#1{equation~\ref{#1}}

\def\1{\bm{1}}

\DeclareMathAlphabet{\mathsfit}{\encodingdefault}{\sfdefault}{m}{sl}
\SetMathAlphabet{\mathsfit}{bold}{\encodingdefault}{\sfdefault}{bx}{n}

\DeclareMathOperator*{\argmax}{arg\,max}

\usepackage{url}

\definecolor{customblue}{RGB}{61,150,209}
\usepackage{wrapfig}
\usepackage[colorlinks=true, citecolor=customblue, linkcolor=customblue, urlcolor=customblue]{hyperref} 
\usepackage{tocbibind} 
\usepackage{titletoc}  

\usepackage[utf8]{inputenc} 
\usepackage[T1]{fontenc}    
\usepackage{booktabs}       
\usepackage{amsfonts}       
\usepackage{nicefrac}       
\usepackage{microtype}      
\usepackage{xcolor}         
\usepackage{amsmath} 
\usepackage{cleveref}
\usepackage{graphicx}
\usepackage{tcolorbox}
\usepackage{amssymb}
\usepackage{caption}  
\usepackage{lipsum}
\usepackage{algorithm}
\usepackage[noend]{algpseudocode} 
\usepackage{subcaption} 
\usepackage[normalem]{ulem}
\usepackage{colortbl}

\tcbset{
  downbox/.style={ 
    colback=green!10,
    colframe=green!10,
    boxrule=0pt,
    boxsep=1pt,
    left=1pt,
    right=1pt,
    top=0pt,
    bottom=0pt,
    baseline=3pt
  },
  upbox/.style={   
    colback=red!10,
    colframe=red!10,
    boxrule=0pt,
    boxsep=1pt,
    left=1pt,
    right=1pt,
    top=0pt,
    bottom=0pt,
    baseline=3pt
  }
}

\newcommand{\upres}[1]{\smash{\tcbox[downbox]{\ensuremath{\uparrow\!#1}}}} 

\usepackage{siunitx}     
\usepackage{array}       

\usepackage{arydshln}
\usepackage{epigraph}
\usepackage{multirow}

\usepackage{enumitem}
\newenvironment{itemize*}%
 {\leftmargini=20pt\begin{itemize}%
  \setlength{\itemsep}{3pt}%
  \setlength{\parskip}{0pt}%
  }%
 {\end{itemize}} 
\newenvironment{enumerate*}%
 {\begin{enumerate}%
  \setlength{\itemsep}{0pt}%
  \setlength{\parskip}{0pt}}%
 {\end{enumerate}}

\definecolor{SAorange}{HTML}{D97706} 
\definecolor{SAforest}{HTML}{1B5E20}  
\definecolor{SAblue}{HTML}{00588B}   

\definecolor{darkgreen}{HTML}{2D8659}

\newcommand{\uncertname}{\textcolor{SAorange}{\textbf{Uncertainty Estimator}}}
\newcommand{\synthname}{\textcolor{SAforest}{\textbf{Data Synthesis Function}}}
\newcommand{\ittname}{\textcolor{SAblue}{\textbf{Test-Time Fine-tuning}}}

\newcommand{\Hunc}{\textcolor{SAorange}{$\mathcal{\textbf{H}}$}}
\newcommand{\Gsyn}{\textcolor{SAforest}{$\mathcal{\textbf{G}}$}}
\newcommand{\Titt}{\textcolor{SAblue}{$\mathcal{\textbf{T}}$}}

\NewDocumentCommand{\heng}
{ mO{} }{\textcolor{red}{\textsuperscript{\textit{Heng}}\textsf{\textbf{\small[#1]}}}}
\NewDocumentCommand{\cheng}
{ mO{} }{\textcolor{orange}{\textsuperscript{\textit{Cheng}}\textsf{\textbf{\small[#1]}}}}

\title{Self-Improving LLM Agents at Test-Time}

\author{Emre Can Acikgoz, \,Cheng Qian, \,Heng Ji, \,Dilek Hakkani-T\"ur, \,Gokhan Tur \\
University of Illinois Urbana-Champaign\\
\texttt{\small{\{acikgoz2,\,chengq9,\,gokhan\}@illinois.edu}}
}

\iclrfinalcopy 
\begin{document}

\maketitle

\begin{abstract}

One paradigm of language model (LM) fine-tuning relies on creating large training datasets, under the assumption that high quantity and diversity will enable models to generalize to novel tasks after post‑training. 
In practice, gathering large sets of data is inefficient, and training on them is prohibitively expensive; worse, there is no guarantee that the resulting model will handle complex scenarios or generalize better.
Moreover, existing techniques rarely assess whether a training sample provides novel information or is redundant with the knowledge already acquired by the model, resulting in unnecessary costs.
In this paper, we explore a new test-time self-improvement method to create more effective and generalizable agentic LMs \textit{on-the-fly}. The proposed algorithm can be summarized in three steps: (i) first it identifies the samples that the model struggles with by using an uncertainty function (self-awareness), (ii) then generates similar examples from the detected uncertain samples (self-data augmentation), and (iii) uses these newly generated samples at test-time fine-tuning (self-improvement). 
We study two variants of this approach: \emph{Test-Time Self-Improvement} (TT-SI), where the same model generates additional training examples from its own uncertain cases and then learns from them, and contrast this approach with \emph{Test-Time Distillation} (TT-D), where a stronger model generates similar examples for those same uncertain cases, enabling the student to adapt using distilled supervision.
Empirical evaluations across different agent benchmarks demonstrate that TT-SI improves the performance with +5.48\% absolute accuracy gain on average across all benchmarks and surpasses other standard learning methods, yet using 68$\times$ less training samples.
TT-D further enhances the performance on challenging scenarios requiring diverse training signals.
Our findings highlight the promise of TT-SI with limitations in current learning frameworks regarding cost and generalization, demonstrating the potential of self-improvement algorithms at test-time as a new paradigm for building more capable agents toward self-evolution. 

\end{abstract}

\begin{figure}[!htb]
    \centering 
    \includegraphics[width=\linewidth]{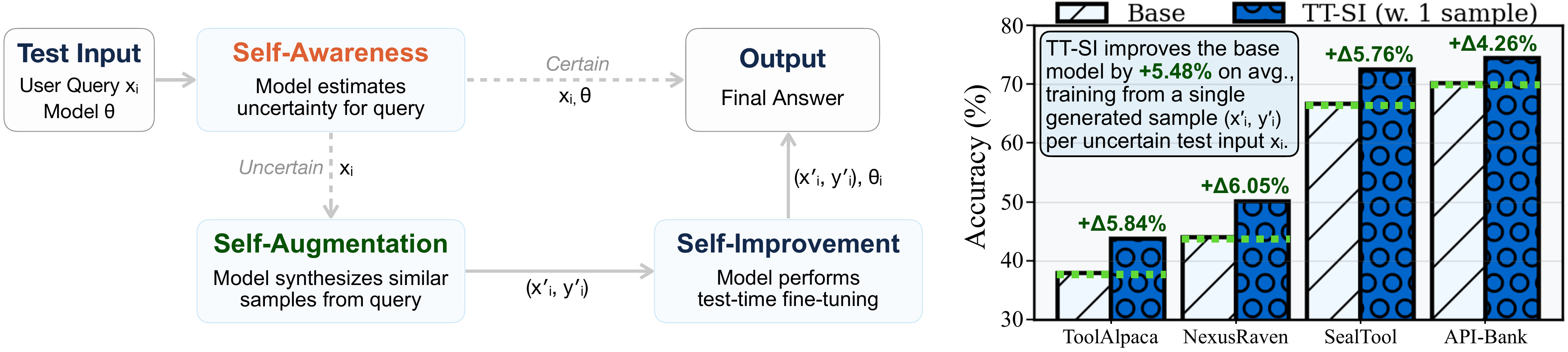}
    \vspace{-4mm}
        \caption{\textbf{Overview of the Test-Time Self-Improvement (TT-SI) framework.}  \textbf{(Left)} TT-SI enables \textit{on-the-fly} adaptation by targeting uncertain test instances during inference.  It consists of three steps: \textbf{(1) Self-Awareness:} An \uncertname\ (\Hunc) identifies challenging samples. \textbf{(2) Self-Data Augmentation:} For each identified uncertain sample, one similar variant is automatically generated using \synthname\ (\Gsyn). \textbf{(3) Self-Improvement:} \ittname\ (\Titt) applies a lightweight update using only \textit{one generated training instance per case}. \textbf{(Right)} $\Delta$-accuracy gains of TT-SI over the prompting baseline at test-time. TT-SI improves the baseline by +5.48\% on average across ToolAlpaca (+5.84\%), NexusRaven (+6.05\%), SealTool (+5.76\%), and API-Bank (+4.26\%).}
    \label{fig:iclr-first}
\end{figure}

\clearpage
\section{Introduction}

Recent progress in language model (LM) post-training has shown promising results across a wide range of tasks~\citep{kumar2025llmposttraining} by equipping these models with explicit knowledge~\citep{grattafiori2024llama, yang2025qwen3}, reasoning~\citep{zelikman2022star, guo2025deepseekr1}, and agentic capabilities~\citep{zeng2024agenttuning, chen2024agentflan}. 
These systems are typically trained to approximate an unknown mapping $\mathcal{F}_{\theta} : \mathcal{X} \rightarrow \mathcal{Y}$ from large‑scale collections of input–output pairs $(x_i, y_i)$, where $\mathcal{X}$ denotes the inputs and $\mathcal{Y}$ denotes their corresponding desired targets. In this approach, a single function $F_{\theta}$ attempts to cover all the relevant knowledge and generalization capability from a single dataset $X$, implicitly assuming that the dataset has sufficient quality, diversity, and scale to effectively learn diverse tasks. 
However, this learning paradigm can remain narrow and inefficient compared to actual human learning~\citep{mitchell2018nel}. 

In contrast, humans take advantage of their background experience (similar to the pretraining stage of LMs) and exhibit remarkable efficiency during learning, often guided by self-regulated learning principles~\citep{zimmerman2002selfregulated} where individuals actively seek and learn from informative demonstrations~\citep{nelson1990metamemory}.
For example, consider a student who is preparing for a college entrance exam after years of coursework. Engaging in metacognitive reflection~\citep{flavell1979metacognition}, the student can either broadly practice questions on various topics (e.g., algebra, history, chemistry) or strategically identify gaps in their knowledge (\textbf{self-awareness}), collect targeted questions addressing these specific deficiencies (\textbf{self-data augmentation}), and practice them repeatedly to learn (\textbf{self-improvement}).  
Clearly, the second strategy is more effective and explicitly improves the required knowledge (see Appendix~\ref{supplement:sport-example} for other examples). 

The same inefficiency is evident in the standard LM agent fine-tuning paradigms, which train the agentic models to inductively learn general rules from training data to be applied to new, unseen test instances such as tool use or other complex agentic tasks.
It involves gathering large-scale training datasets~\citep{ouyang2022rlhf, wang2023selfinstruct, zeng2024agenttuning} (either human-curated or LLM-synthesized) and fine-tuning models on these datasets~\citep{grattafiori2024llama, zeng2024agenttuning, acikgoz2025can}. 
However, constructing these datasets is costly, often requiring days to weeks of computation and manual labor, and still provides no guarantee of effective performance and generalization after fine-tuning. 
Moreover, this approach implicitly assumes that models must process every sample, without considering if certain examples are redundant or already known by the LM.
Based on these deficiencies, a key open question is \textit{whether models can be trained to acquire new skills more efficiently, without relying on exhaustive datasets or processing large amounts of redundant information}.

Motivated by local and transductive learning~\citep{bottou_and_vapnik1992locallearning, joachims1999transductive} with recent advances in test-time fine-tuning~\citep{akyurek2025tttnlp}, we investigate a simple, yet powerful, instance-specific self-improvement algorithm that adapts agents \textit{on-the-fly} to each downstream task at test-time (\Cref{fig:iclr-first}). The proposed algorithm first identifies the most informative and challenging samples while discarding mastered or redundant ones, guided by the designed \uncertname\ (\Hunc), which reflects \textit{self-awareness}. For each retained “necessary” test instance, the model synthesizes a set of distributionally similar samples with \synthname\ (\Gsyn) as \textit{self-data augmentation} and performs temporary gradient updates with \ittname\ (\Titt) through \textit{self-improvement} on these instances. We explore two different variants of our approach: Test-Time Self-Improvement (TT-SI), where the model trains on self-generated samples using parameter efficient fine-tuning techniques (PEFT)~\citep{hu2022lora}, and Test-Time Distillation (TT-D) where adaptation is guided by supervision from samples synthesized by a more capable teacher model. 

We demonstrate that \textit{test-time self-improvement} enables agents to adapt on-the-fly by leveraging their own uncertain predictions. With only a \emph{single synthesized training instance} per test case, TT-SI shows consistent absolute accuracy gains across four challenging agent benchmarks: +5.84\% on ToolAlpaca +6.05\% on NexusRaven, +5.76\% on SealTool, and +4.26\% on API-Bank. These improvements highlight that even minimal, uncertainty-guided adaptation can substantially boost performance during inference.  Moreover, TT-D further extends these gains in complex, context-heavy scenarios (e.g., multi-turn conversations). Compared to standard supervised fine-tuning (SFT), TT-SI surpasses accuracy on SealTool while using 68$\times$ fewer samples, underscoring efficiency without compromising effectiveness. We find that, when training is infeasible, TT-SI with in-context learning (ICL) offers a fast, training-free alternative, outperforming other standard learning methods in similar conditions.

Concretely, our main findings and contributions can be summarized as follows:
\begin{itemize}[leftmargin=*]
    \itemsep -0.4ex
    \item We propose a three-stage algorithm for \textit{test-time self-improvement}, motivated by human learning theories: (i) identify uncertain samples via a novel uncertainty estimator, (ii) generate new training instances similar to these samples, and (iii) update the model online.
    \item We conduct a systematic empirical study of two variants, TT-SI and TT-D, analyzing key components such as the impact of uncertain samples, learning method at test time, scaling of generated samples, and other parameter effects.
    \item We validate that agentic LMs can self-improve during inference, even from a single training instance, and show that our framework outperforms standard inductive learning approaches, achieving significant gains with orders-of-magnitude less compute through both test-time ICL and test-time fine-tuning.
\end{itemize}

Overall, our work pioneers a novel self-improvement algorithm for agent learning, inspired by human-like lifelong adaptation, seamlessly integrating self-awareness, targeted self-generated data, and iterative self-training to enable continuous self-improvement. We propose that with an optimal uncertainty estimator to identify weaknesses, precise data synthesis to address them, and focused iterative training, agents can continually advance toward mastering increasingly complex and diverse tasks. Thus, this study opens new research directions along these three interconnected paths, \textbf{rethinking how LM agents learn, adapt, and generalize.}

\section{Preliminaries}

\subsection{Fundamental Issues in Inductive Fine-Tuning}
The standard post-training paradigm separates training and testing: models are trained by \textit{inductively} extracting generalizable patterns from data and subsequently evaluated on new, possibly unseen examples~\citep{vapnik1995nature, lecun2015deep, zhang2024sftsurvey}. Current approaches for training LMs largely follow this paradigm, relying on large-scale post-training datasets. Formally, these datasets are denoted as $\mathcal{D}_{\text{train}} = \{(x_i, y_i)\}_{i=1}^{N}$, consist of $N$ samples assumed to be independently and identically distributed (i.i.d.) according to a \textit{training} distribution $\mathcal{P}_{\text{train}}(\mathcal{X}, \mathcal{Y})$. 

Here, $x_i \in \mathcal{X}$ represents an input (e.g., a task query) and $y_i \in \mathcal{Y}$ is its corresponding desired output (e.g., a sequence of actions for an agent). The objective is to find the parameters $\theta$ of a mapping function $\mathcal{F}_{\theta}: \mathcal{X} \rightarrow \mathcal{Y}$, representing the agent, that minimize the empirical risk on the training data:
$\hat{\mathcal{L}}_{\text{train}}(\theta) = \frac{1}{N} \sum_{i=1}^{N} \ell(\mathcal{F}_{\theta}(x_i), y_i)$,
where $\ell$ is a predefined loss function. The foundational assumption is that if $N$ is sufficiently large and $\mathcal{D}_{\text{train}}$ is diverse enough, the learned model $\mathcal{F}_{\theta}$ will generalize effectively to new, unseen inputs $x$ drawn from a \textit{test} distribution $\mathcal{P}_{\text{test}}(\mathcal{X})$. However, this prevailing paradigm is beset by several fundamental issues: 
\begin{itemize}[leftmargin=*]
\itemsep -0.4ex
\item \textbf{Distributional Shift:} Test distributions $\mathcal{P}_{\text{test}}$ often differ from the training distribution $\mathcal{P}_{\text{train}}$ (i.e., $\mathcal{P}_{\text{test}} \neq \mathcal{P}_{\text{train}}$). 
This means the empirical risk $\hat{\mathcal{L}}_{\text{train}}(\theta)$ provides a misleading picture of the true test risk $\mathcal{L}_{\text{test}}(\theta) = \mathbb{E}_{(x,y)\sim\mathcal{P}_{\text{test}}}[\ell(\mathcal{F}_{\theta}(x),y)]$, which in turn impairs the model's generalization to novel or complex scenarios~\citep{liu2021oodsurvey}.
\item \textbf{Computation Cost:} The reliance on extremely large training datasets $\mathcal{D}_{\text{train}}$ (with $N \gg 10^4$ samples) leads to substantial annotation and computational costs, both scaling with $N$, rendering agent development prohibitively expensive~\citep{mirzasoleiman2020coresetsdata, mindermann2022prioritized}. 
\item \textbf{Redundancy and Inefficient Use of Information:} Treating all $N$ training samples $(x_i, y_i)$ in $\mathcal{D}_{\text{train}}$ as equally informative is inefficient, as the number of truly effective samples $N_{\text{eff}}$ is often much smaller than $N$ (i.e., $N_{\text{eff}} \ll N$)~\citep{zhou2023lima}. Processing redundant or already mastered examples wastes computational effort and can even restrict generalization, particularly for inputs from the long tail of the data distribution or adversarial examples, which current agents struggle with~\citep{Settles2009ActiveLL, sorscher2022beyondpruning}.
\item \textbf{Catastrophic Forgetting and Model Churn:} Standard fine-tuning for LMs often suffer from catastrophic forgetting~\citep{luo2023catastrophicsurvey}, where fine-tuning a model on a new task inadvertently degrades its performance on previously acquired skills.  Moreover, the rapid release of new and more capable LLMs~\citep{grattafiori2024llama, yang2025qwen3} necessitates a continuous and costly re-training cycle, as the entire fine-tuning process must be repeated on $\mathcal{D}_{\text{train}}$ to leverage the increased knowledge and reasoning abilities of each new base model on the downstream task.

\end{itemize}

These limitations motivate a new post-training paradigm grounded in \textit{transductive} and \textit{local learning} principles, which adapts the model on-the-fly by identifying and training only on the most informative samples drawn from the test distribution $\mathcal{P}_{\text{test}}$.

\subsection{Test-Time Training}
Test-time training (TTT) performs small, ephemeral parameter updates during inference, conditioning the model on the current input and thus partially collapsing the train–test boundary \citep{sun2023tttphd}. The idea traces to local and transductive learning, where hypotheses are adapted after observing test inputs \citep{bottou_and_vapnik1992locallearning,joachims1999transductive}. In deep learning, \citet{sun2020tttcv} showed that a simple self-supervised TTT objective can improve the robustness of image classifiers under distribution shift. In LLMs, TTT is comparatively nascent: \citet{hardt2024tttknn} fine-tune on retrieved nearest neighbors to reduce perplexity, and SIFT \citep{hubotter2025sift} actively selects diverse, informative neighbors to limit redundancy. Closest to our setting, \citet{akyurek2025tttnlp} apply rule-based linear transformations to in-context test examples in ARC to get additional test-time training data. However, these approaches either target perplexity rather than general reasoning tasks, assume access to high-quality neighbors, or in-context exemplars. Our work instead selects informative test instances, generates and filters training signals on-the-fly, yielding improvements on challenging agent benchmarks. To the best of our knowledge, this is the first language generation–based test-time fine-tuning method applied to LLM-based agents. Further details on prior work in LLM and agent post-training, and how our work differs, are provided in Appendix \Cref{supplement:agent-related-work}.

\subsection{Self-Improvement in LLMs and How it Works}

Recent studies suggest that LLMs can \textit{self-improve} their capabilities~\citep{huang2023llmscanimprove, wang2023selfinstruct, chen2024teachingselfdebug, pang2024language_self_rl, yuan2024selfreward, shafayat2025canselftrain, huang2025selfimprovement}, i.e., they can refine their own output distribution using internal signals derived from their own parameters, without relying on external supervision~\citep{xie2020selfcvpr, He2020Revisitingnsg, huang2023llmscanimprove, huang2025selfimprovement}. At first glance, this appears paradoxical: how can a model improve its performance if no new information is introduced? The key insight lies in the hypothesis that LLMs contain \textit{hidden knowledge}~\citep{hinton2015distilling}, latent representations within their weights that are not fully accessible through standard inference. \cite{huang2025selfimprovement} suggests self-improvement emerges through a \textit{sharpening mechanism}, where the model iteratively refines its output distribution to favor high-confidence predictions that align with internal self-evaluation criteria, effectively surfacing this hidden knowledge. This process can be framed as \textbf{distribution sharpening}. 

Formally, let $\theta_0$ denote the parameters of a base model $\mathcal{M}$. For a test input $x_i \in \mathcal{D}_{\text{test}}$, the model induces a conditional distribution $\mathcal{M}_{\theta_0}(y|x_i)$ over possible responses $y$. Self-improvement methods aim to adapt $\theta_0$ such that the updated parameters $\theta_i$ favor responses that maximize an internally defined self-reward $r_{\text{self}}$:
\begin{equation}
   \theta_i \approx \argmax_{\theta} \, r_{\text{self}}(y|x_i,\theta), \quad y \sim \mathcal{M}_{\theta}( \cdot | x_i).
   \label{eq:sharp}
\end{equation}
Here, $r_{\text{self}}$ is not explicitly optimized but acts as an \textit{implicit intrinsic reward}, induced by the model’s own objective and activated during adaptation~\citep{agarwal2025unreasonableentorpy, shafayat2025canselftrain, zuo2025ttrl}. Overall, this process tilts the distribution toward more certain, high-reward outputs, amplifying the model's inherent strengths. Self-improvement, therefore, is not about creating knowledge ex nihilo, but rather about designing algorithms to elicit and amplify this hidden latent knowledge. 

Building on this foundation, our test-time self-improvement (TT-SI) algorithm operationalizes the sharpening mechanism by learning sample-specific temporary parameters $\theta_i$ during inference: it first detects uncertain inputs $x_i$ via \uncertname\ (\Hunc),  targeting cases where latent knowledge is most accessible yet underutilized, then it synthesizes similar training instances from these uncertain examples with \synthname\ (\Gsyn), and finally performs targeted test-time fine-tuning using \ittname\ (\Titt) on these samples, thereby maximizing $r_{\text{self}}$ only where adaptation yields the highest marginal benefit.

\section{Method}
\label{sec:method}

We introduce a test-time self-improvement framework designed to enable agents to learn from challenging instances on-the-fly by integrating three key components, as shown in \Cref{alg:uncertainty-inference}:
\begin{itemize}[leftmargin=*]
\itemsep -0.4ex
    \item \textbf{Self-Awareness:} \textbf{\uncertname\ (\Hunc)} identifies inputs $x_i$ at inference-time which the agent is uncertain on, ensuring adaptation focuses only on informative, challenging cases (\Cref{subsec:uncertain-sample-selection}).
    \item \textbf{Self-Augmentation:} \textbf{\synthname\ (\Gsyn)} generates a set of $K$ new samples ($\mathcal{D}'_i$) that are closely related synthetic examples, generated based on the uncertain input $x_i$ (\Cref{subsec:synthesize-similar-samples}).
    \item \textbf{Self-Learning:} \textbf{\ittname\ (\Titt)} temporarily updates the agent's parameters ($\theta$) on the targeted synthetic data ($\mathcal{D}'_i$) (\Cref{subsec:inference_time_finetuning}).
\end{itemize}
In the following subsections, we detail each component one by one, first by providing formal definitions followed by their algorithmic specifics.

\begin{algorithm}[t!]
\caption{Test-Time Self-Improvement Framework}
\label{alg:uncertainty-inference}
\begin{algorithmic}[1]
\Require Test dataset $\mathcal{D}_{\text{test}}$, model $\mathcal{M}$, data generation prompt $\mathcal{P}$, temporary dataset size $K$, initial model parameters $\theta_0$

\For{each $x_i \in \mathcal{D}_{\text{test}}$}

    \State \textbf{Step 1:} \uncertname\ (\Hunc)
    \State \textbf{Compute} uncertainty (softmax-difference):

    \State \quad $\ell_n = -\log P_{\mathcal{M}}(a_n|x_i), \quad \forall a_n$ \Comment{Negative Log-Likelihood (NLL) for candidate action}
    \State \quad $p_n = \frac{\exp(\ell_n - \max_j \ell_j)}{\sum_{k}\exp(\ell_k - \max_j \ell_j)}$ \Comment{Apply Relative Softmax Scoring (RSS) normalization}

    \State \quad $u(x_i) = p^{(1)} - p^{(2)}$ \Comment{Highest minus second-highest RSS scores}

    \State \textbf{Step 2:} \synthname\ (\Gsyn)
    \State \textbf{if} $u(x_i) < \tau$ \textbf{then} \Comment{Check uncertainty}

    \State \quad \textbf{Generate} $K$ synthetic samples using LLM:
    \State \quad \quad $\mathcal{D}_i \leftarrow \mathcal{L}_{\text{gen}}(x_i, K)$ \quad \Comment{\Cref{eq:synthesis-function}}

    \State \textbf{Step 3:} \ittname\ (\Titt)
    \State \quad \textbf{Learn} temporary model parameters $\theta_i^*$ via LoRA:
    \State \quad \quad $\theta_i^* \leftarrow \arg\min_{\theta_0} \sum_{(x',y')\in \mathcal{D}_i} \ell(\mathcal{M}(x'; \theta_0), y')$ \quad \Comment{\Cref{eq:adaptation_loss}}

    \State \quad \textbf{Perform} inference with adapted parameters $\theta^*_i$:
    \State \quad \quad $\hat{y}_i \leftarrow \mathcal{M}(x_i; \theta_i^*)$

    \State \quad \textbf{Reset} model parameters:
    \State \quad \quad $\theta_i^* \rightarrow \theta_0$ \quad \Comment{Restore original parameters}

    \State \textbf{else}
    \State \quad \textbf{Perform} inference directly:
    \State \quad \quad $\hat{y}_i \leftarrow \mathcal{M}(x_i; \theta_0)$

    \State \textbf{end if}

\EndFor

\end{algorithmic}
\end{algorithm}

\subsection{Self-Aware Sample Selection at Test Time}
\label{subsec:uncertain-sample-selection}

This section details our approach for identifying and selecting data samples for which the model $\mathcal{M}$ exhibits high uncertainty during inference. 
We posit that such samples are more likely to be challenging or error-prone, and are thus particularly informative for further learning.

\noindent\textbf{Definition\;\;} Given a task with inputs $x_i$, we define \uncertname\ (\Hunc) that estimates the model’s confidence score ($\mathcal{C}$) for each candidate action ${a_1, ..., a_n}\in\mathcal{A}$ available to the model $\mathcal{M}$ in its environment (e.g., available API calls). For each input $x_i$ and candidate action $a_n$, the confidence is computed as:
\begin{equation}
\mathcal{C}_{i} = \textcolor{SAorange}{\mathcal{\textbf{H}}}(x_i, a_n, \mathcal{M})
\end{equation}
This estimation is performed without access to ground-truth labels $y_i$, ensuring fairness and applicability during inference. A sample $x_i$ is deemed \textbf{uncertain} if $\mathcal{C}{_i} < \tau$ for a user-defined confidence threshold $\tau$. By filtering out high-confidence (i.e., certain) instances, this uncertainty estimation step focuses computational and learning resources for the most informative and challenging questions, thereby enhancing both efficiency and quality.

\noindent\textbf{Selecting Uncertain Samples\;\;} 
To systematically identify uncertain samples, we implement a \emph{margin-based confidence estimator} using the likelihood distribution generated by the model $\mathcal{M}$ for a given input $x_i$. Given a set of available actions ${a_1, a_2, \dots, a_N}$, we first compute the negative log-likelihood (NLL) for each action as:
\begin{equation}
\mathrm{NLL}(a_n | x_i) = -\log P_{\mathcal{M}}(a_n | x_i), \quad \forall n \in {1, 2, \dots, N}.
\label{eq:confidence}
\end{equation}
However, raw NLL scores are not directly interpretable due to their unbounded nature, limiting their utility in precisely quantifying uncertainty. To address this issue, we apply a \emph{Relative Softmax Scoring (RSS)} mechanism, which transforms these scores into a normalized and interpretable confidence distribution:
\begin{equation}
p^n = \frac{\exp(\ell_n - \max_j \ell_j)}{\sum_{k=1}^{N}\exp(\ell_k - \max_j \ell_j)}, \quad\text{where}\quad \ell_n = -\mathrm{NLL}(a_n \mid x_i).
\label{eq:rss}
\end{equation}
Here, $p^n$ is the RSS confidence score for action $a_n$, and $\ell_n$ denotes the negative log-likelihood score corresponding to $a_n$. The $\max_j \ell_j$ term represents the maximum NLL score among all candidate actions, serving as a numerical stabilizer. To quantify prediction uncertainty, we compute the difference between the highest and second-highest RSS scores, termed the \emph{softmax-difference}. Formally, uncertainty for input $x_i$ is defined as:
\begin{equation}
    u(x_i) = p^{(1)} - p^{(2)},
\label{eq:diff}
\end{equation}
where $p^{(1)}$ and $p^{(2)}$ denote the highest and second-highest RSS scores, respectively. Finally, using a user-defined threshold $\tau$, we select samples exhibiting high uncertainty ($u(x_i) < \tau$), which ensures that subsequent adaptation or analysis efforts are focused on the most ambiguous instances, where the model is likely to benefit most from further information or refinement. 
\subsection{Data Generation Strategies}
\label{subsec:synthesize-similar-samples}
Once an individual input sample $x_i$ is identified as exhibiting high uncertainty by the model $\mathcal{M}$ (as per the criteria in \Cref{subsec:uncertain-sample-selection}), our approach triggers an immediate data synthesis process with \synthname\ (\Gsyn). This section details the methodology for generating new, relevant training data specifically for the uncertain instance at hand. The core idea is to create a focused, temporary dataset on-the-fly, enabling rapid, localized adaptation of the model to address the specific query it found challenging.

\subsubsection{Data Synthesis Method}

\noindent\textbf{Definition\;\;} 
When an input sample $x_i$ (without ground-truth labels) is processed during inference and flagged as uncertain by \Hunc, we trigger the synthesis of $K$ new training examples together with the corresponding labels. \synthname\ (\Gsyn) is invoked for this specific uncertain input $x_i$, producing a set of $K$ new input-output pairs following the provided instructions (\Cref{tab:data-generation-prompt}). These synthetic examples, $(x'_{ij}, y'_{ij})_{j=1}^{K}$, are aimed to be semantically similar to the original uncertain sample $x_i$ while introducing slight variations. In practice, $x_i$ serves as a seed example in the prompt, guiding the generation of $K$ new synthetic training pairs $(x', y')$ that resemble the original input but expand the training signal~\citep{wang2023selfinstruct}.

The \Gsyn\ is invoked for this specific uncertain input $x_i$, producing a set of $K$ novel input-output pairs following the provided prompt of instructions ($\mathcal{P}$):
\begin{equation}
\textcolor{SAforest}{\mathcal{\textbf{G}}}: x_i \rightarrow \{(x'_{ij}, y'_{ij})\}_{j=1}^{K}
\label{eq:synthesis-function}
\end{equation}
Here, $K$ is a user-defined hyperparameter dictating the volume of synthetic data generated for the current uncertain instance $x_i$. Each generated pair $(x'{ij}, y'{ij})$ aims to be a plausible instance from the underlying data distribution $P(x, y)$ relevant to $x_i$, specifically targeting the model's region of uncertainty around this input. These $K$ generated pairs immediately form a temporary, query-specific dataset $\mathcal{D}_i$:
\begin{equation}
\mathcal{D}_{i} = \{(x'_{ij}, y'_{ij})\}_{j=1}^K
\label{eq:temporary-dataset}
\end{equation}

This dataset $\mathcal{D}_i$ is then used for a localized adaptation of the model parameters $\theta$ before processing subsequent input samples with \ittname\ (\Titt) as the next step described in the next section (\Cref{subsec:inference_time_finetuning}). This iterative process of detection, synthesis, and adaptation is performed for each sample identified as uncertain.

\noindent\textbf{Generating Samples\;\;} The implementation of the synthesis function \Gsyn\ (\Cref{eq:synthesis-function}), which is triggered for each uncertain sample $x_i$, employs the agent itself for data synthesis ($\mathcal{L}_{\text{gen}}$) as \textit{self-augmentation}. For each generation instance, $\mathcal{L}_{\text{gen}}$ is provided with a carefully hand-crafted prompt $\mathcal{P}$ (See \Cref{tab:data-generation-prompt}), the uncertain input $x_i$ serving as the direct seed (critically, without its corresponding label $y_i$), and a specified number of samples $K$ to generate. The model then produces $K$ new input-output pairs, denoted as $\{(x'_{ij}, y'_{ij})\}_{j=1}^{K}$. This seed-based generation process, inspired by self-instruction methodologies~\citep{wang2023selfinstruct}, guides $\mathcal{L}_{\text{gen}}$ to produce variants that maintain the core semantic meaning and task relevance of $x_i$ while introducing controlled surface-level variations. By \textit{synthesizing data in this on-the-fly} manner for each uncertain instance, we facilitate targeted and timely model adaptation, aiming to improve performance on precisely the types of queries the model struggles with, as they are encountered.

\subsection{Test-Time Fine-Tuning}
\label{subsec:inference_time_finetuning}
Test-Time Fine-Tuning enables parametric models to update their weights temporarily during inference~\citep{sun2023tttphd}, yet this paradigm is largely unexplored for LLMs~\citep{akyurek2025tttnlp} and, to the best of our knowledge, has not been applied to agentic tasks. Building on our previous steps, uncertainty detection (\Cref{subsec:uncertain-sample-selection}) and targeted data synthesis (\Cref{subsec:synthesize-similar-samples}), we now use \ittname\ (\Titt) to adapt model $\mathcal{M}$ during inference, using the generated samples $\mathcal{D}_i$ for each uncertain test query $x_i$.

\noindent\textbf{Definition\;\;}  Once we got $\mathcal{D}_i$ with \Cref{eq:synthesis-function}, we optimize initial parameters ($\theta_0$) to minimize the loss function $\mathcal{L}(\mathcal{D}_i;\theta_0)$, producing temporarily updated parameters $\theta_i$ for the target task prediction. Importantly, after generating predictions, the model is restored to the original parameters $\theta_0$ for the next iteration using sample $x_{i+1}$, thereby creating a specialized prediction model for each \textit{out-of-distribution} sample without permanently altering the base model.

\noindent\textbf{Test-Time Fine-Tuning\;\;} The primary goal of inference-time adaptation is to temporarily adjust the model $\mathcal{M}$'s parameters ($\theta$) to better handle the current uncertain sample $x_i$. This is achieved by fine-tuning the model on the newly synthesized dataset $\mathcal{D}_i = \{(x'_{ij}, y'_{ij})\}_{j=1}^{K}$. The adaptation involves minimizing a task-specific loss function $\mathcal{L}_{\text{task}}$ over the samples in $\mathcal{D}_i$. For a given self-generated sample $(x'_{ij}, y'_{ij}) \in \mathcal{D}_i$, the loss is computed as $\ell(\mathcal{M}(x'_{ij}; \theta), y'_{ij})$, and the objective for adapting parameters $\theta$ using dataset $\mathcal{D}_i$ is:
\begin{equation}
\theta_i^* = \arg\min_{\theta'} \sum_{(x'_{ij}, y'_{ij}) \in \mathcal{D}_i} \ell(\mathcal{M}(x'_{ij}; \theta'), y'_{ij})
\label{eq:adaptation_loss}
\end{equation}
where $\theta_i^*$ represents the adapted parameters for the context of $x_i$. We employ Low-Rank Adaptation (LoRA)~\citep{hu2022lora} to ensure computational efficiency during inference-time updates.

\section{Results}
\label{sec:results}

\noindent\textbf{Experimental Protocol\;\;} We evaluate our approach on four complementary agent benchmarks. NexusRaven~\citep{srinivasan2023nexusraven} is a function-calling benchmark that tests an agent’s ability to execute single, nested, and parallel function calls of varying complexity. SealTool~\citep{wu2024sealtools} is a self-instruct dataset for tool learning, measuring precision in tool selection, adherence to output formats, and adaptability across diverse scenarios. API-Bank~\citep{li2023api} evaluates multi-turn user–agent dialogues, requiring agents to track conversational state, make informed tool calls at each turn, and handle realistic conditions such as noisy or incomplete inputs. Finally, ToolAlpaca~\citep{tang2023toolalpaca} is designed for tool-learning that employs synthetic data generation to create 271 tool-calling instances across 50 different categories. 
We use \texttt{Qwen2.5-1.5B-Instruct} for main experiments, as its strong performance and small size allow efficient use of limited hardware and demonstrate the potential of compact agentic models~\citep{belcak2025smallagents}. To examine scaling and architectural variations, we further include \texttt{Qwen2.5-7B-Instruct} ablations.
All models are trained with PEFT using LoRA~\citep{hu2022lora} on a single NVIDIA A40 GPU. Because our method often follows a small-sample regime (e.g., single-sample training), higher deviation is expected; thus, all experiments, including baselines, are repeated five times with different sample trainings, seeds, and reported as averages.
Additional details are provided in Appendix \Cref{supplement:reproductibility}.

\begin{table}[!htb]
\centering
\small
\renewcommand{\arraystretch}{1.}
\resizebox{\textwidth}{!}{%
\begin{tabular}{lccccccc}
\toprule
\textbf{Inference} & \textbf{Method} & \textbf{NexusRaven} & \textbf{SealTool} & \textbf{API-Bank} & \textbf{ToolAlpaca} & \textbf{Avg.} & $\Delta\%$ \\
\midrule
\multirow{3}{*}{Input/Output}
& w/o TT-SI {\scriptsize(Base)}     & 44.03\tiny{$\pm$1.42} & 66.67\tiny{$\pm$2.39}  & 70.08\tiny{$\pm$0.82}  & 37.86\tiny{$\pm$0.97}  & 54.66 & -- \\
& w.\ TT-SI                         & 50.08\tiny{$\pm$0.47} & 72.43\tiny{$\pm$0.75}  & 74.34\tiny{$\pm$1.89}  & 43.70\tiny{$\pm$0.68}  & 60.14 & \upres{5.48} \\
& w.\ TT-D                          & 52.52\tiny{$\pm$0.65} & 73.92\tiny{$\pm$0.79} & 75.56\tiny{$\pm$0.57} & 42.31\tiny{$\pm$0.87} & 61.08 & \upres{6.42} \\
\midrule
\multirow{3}{*}{Majority Vote}
& w/o TT-SI {\scriptsize(Base)}     & 46.56\tiny{$\pm$1.61}  & 69.73\tiny{$\pm$1.21}  & 73.96\tiny{$\pm$0.75}  & 41.94\tiny{$\pm$0.81}  & 58.05 & -- \\
& w.\ TT-SI                         & 52.20\tiny{$\pm$1.34}  & 72.93\tiny{$\pm$0.59}  & 75.68\tiny{$\pm$0.74}  & 46.79\tiny{$\pm$1.06}  & 61.90 & \upres{3.85} \\
& w.\ TT-D                          & 54.91\tiny{$\pm$0.57} & 75.28\tiny{$\pm$0.86} & 78.12\tiny{$\pm$0.71} & 46.02\tiny{$\pm$0.53} & 63.58 & \upres{5.53} \\
\midrule
\multirow{3}{*}{Pass@5}
& w/o TT-SI {\scriptsize(Base)}  & 59.69\tiny{$\pm$0.51}  & 78.16\tiny{$\pm$0.92}  & 78.67\tiny{$\pm$0.69}  & 46.99\tiny{$\pm$0.53}  & 65.88 & -- \\
& w.\ TT-SI                      & 63.40\tiny{$\pm$0.26}  & 82.32\tiny{$\pm$0.67}  & 81.72\tiny{$\pm$0.58}  & 49.90\tiny{$\pm$0.87}  & 69.34 & \upres{3.46} \\
& w.\ TT-D                       & 65.98\tiny{$\pm$0.57} & 84.78\tiny{$\pm$0.53} & 84.97\tiny{$\pm$0.89} & 52.24\tiny{$\pm$0.82} & 71.99 & \upres{6.11} \\
\bottomrule
\end{tabular}
}
\caption{\textbf{Main Results of TT-SI}. Accuracy results of baseline prompting (w/o TT-SI), TT-SI, and TT-D across four agentic benchmarks under three inference settings: \emph{Input/Output} (direct prediction) and \emph{Majority Vote} (5-sample self-consistency), and \emph{Pass@5} (correct if any of 5). $\Delta$ denotes the average absolute improvement over base model without TT-SI and {\color{green!50!black}$\uparrow$} indicates performance increase.}
\vspace{-5mm}
\label{tab:results-delta}
\end{table}

\subsection{Main Results}
\vspace{-1mm}

\noindent\textbf{Insight 1: Agents can self-improve themselves at test-time even when training on just one sample.}
For our main results, we evaluate TT-SI on four agentic benchmarks: NexusRaven, SealTool, API-Bank, and ToolAlpaca, using three different inference method as direct zero-shot prompting, majority vote over 5 generations, and pass@5 (success if any of 5 generations is correct), as shown in \Cref{tab:results-delta}.
Here we check the effect of TT-SI with SFT by only using one sample generated by \Gsyn\ and identified through \Hunc. 
TT-SI improves the baseline (w/o TT-SI) across all benchmarks, achieving an average absolute gain of 5.48\% for direct inference (54.66\%$\rightarrow$65.62\%), 3.85\% for majority voting (58.05\%$\rightarrow$61.90\%), and 3.46\% for pass@5 (65.88\%$\rightarrow$69.34\%), which shows TT-SI enables agents to self-improve with only one training instance per uncertain case during inference.
Here, TT-SI acts a test-time \emph{sharpening} step where one synthetic sample from an uncertain input re-weights the model’s probability distribution to resolve that uncertainty.
We also examine a variant of TT-SI as test-time distillation (TT-D), where where \Gsyn’s self-generated data is replaced with \texttt{gpt-5-mini} outputs. TT-D further improves over TT-SI by 0.94\% for direct inference, 1.68\% for majority voting, and 2.65\% for pass@5, indicating that higher quality training signals provide modest but consistent additional gains.

\noindent\textbf{Insight 2: TT-SI outperforms inductive SFT with orders of magnitude less data.}  
We compare TT-SI against in-context learning (ICL, 1-shot) and supervised fine-tuning (SFT) on SealTool, which provides an official training split of $\sim$13k samples (\Cref{fig:two-side-by-side}, left). TT-SI with SFT (72.43\%) surpasses all three baselines and exceeds standard inductive SFT (70.20\%) by $2.23\%$ accuracy. Notably, TT-SI achieves this improvement using only 190 uncertain cases (each paired with one synthetic example) rather than the full 13k training set. This corresponds to roughly \textbf{68$\times$ fewer samples}, yet delivers better accuracy, highlighting TT-SI as an effective alternative to conventional learning approaches.

\noindent\textbf{Insight 3: When training is infeasible, test-time self-improvement with ICL offers a fast alternative.}
We extend TT-SI to an ICL setting (\Cref{fig:two-side-by-side}, left), where generated examples are inserted directly into the context of the prompt rather than used for fine-tuning. TT-SI with ICL achieves a slight improvement over the base model (66.37\%$\rightarrow$68.36\%) and even outperforms the standard ICL baseline (67.74\%) that leverages SealTool’s training split. This highlights ICL as a training-free, low-overhead alternative to inductive methods. This improvement is likely to be from enhanced model certainty: TT-SI generates demonstrations that boost the model's confidence in the correct output format and reasoning process, increasing the likelihood of accurate predictions without relying on external training data.

\noindent\textbf{Insight 4: Uncertainty filtering balances accuracy and efficiency.}  
Because TT-SI operates at inference time, efficiency is critical. In our design, \Hunc\ identifies uncertain samples for targeted adaptation, while certain ones are passed directly to the base model. To assess its effect, we also evaluate a variant that treats all test inputs as uncertain (TT-SI w/o \Hunc) in \Cref{fig:two-side-by-side} (left). This achieves only a marginal $+1.04\%$ gain, but requires adapting to all 294 test samples rather than 190 samples (an additional 104 LoRA weights to learn) resulting in higher cost. Thus, the slight accuracy gain is outweighed by the efficiency loss, underscoring the importance of uncertainty filtering for practical test-time adaptation. The estimator’s accuracy is further detailed in \Cref{supplement:uncertainty-estimator}.

\subsection{Ablation Studies and Analysis}
\label{subsec:ablations}

\begin{figure}
    \centering
    \includegraphics[width=\linewidth]{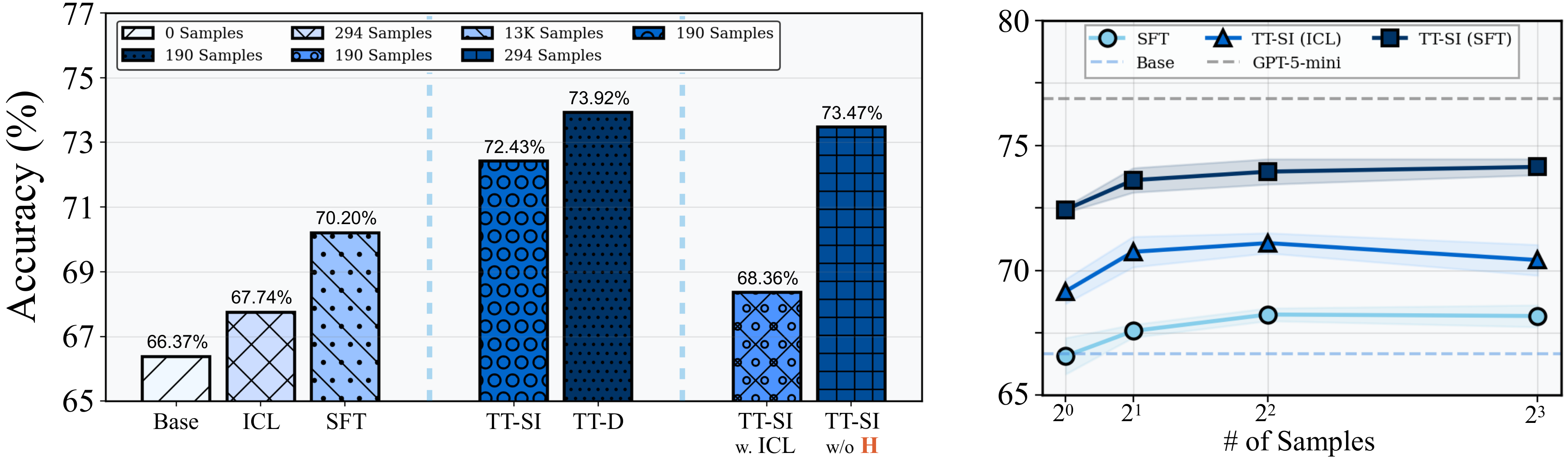}
    \caption{\textbf{Experimental Results on SealTool.} 
\textbf{Left}: Accuracy comparison of TT-SI against standard baselines (left-most) and its variants (middle), including ablations (right-most). \textbf{Right}: Scaling behavior under different adaptation strategies with varying numbers of samples. Shaded regions show variance over five runs; dashed lines denote baseline references.}
    \vspace{-5mm}
    \label{fig:two-side-by-side}
\end{figure}

\noindent\textbf{Insight 5: Data scaling on OOD data highlights the limits of SFT and the strength of TT-SI.}  
We examine data scaling for both standard SFT and TT-SI (\Cref{fig:two-side-by-side}, right). Here, we leverage the state-of-the-art xLAM function-calling dataset~\citep{zhang2025xlam} for SFT as an out-of-distribution (OOD) setting. For each scale (1, 2, 4, 8), we sample five subsets for training and report averages with standard deviations.
TT-SI consistently outperforms SFT across all scales, with improvements growing as more uncertain examples are incorporated, highlighting the importance of uncertainty-guided data and the value of targeted test-time learning. 
Moreover, the training-free variant of TT-SI with ICL also surpasses standard SFT on a strong dataset using the same data amounts per scale, which shows that even without a dedicated training split or fine-tuning, test-time approaches can outperform SFT under same conditions on OOD data.

\noindent\textbf{Insight 6: Optimal $\tau$ improves efficiency with minimal accuracy loss.} 
We investigate the impact of the uncertainty threshold $\tau$ on TT-SI performance and efficiency in \Cref{tab:uncertainty-baselines}.
First, TT-SI improves accuracy regardless of $\tau$, surpassing the base of 66.37\% in all cases.
\begin{wraptable}{r}{0.48\textwidth}
  \vspace{-1mm}
  \centering
  \resizebox{0.48\columnwidth}{!}{
  \begin{tabular}{l c c c c}
        \toprule
        \textbf{$\tau$ / Setting} & \textbf{TPR} & \textbf{FPR} & \textbf{Unc. ($\Delta$\%)} & \textbf{Acc.} \\
        \midrule
        Base          & {\text{--}} & {\text{--}} & {\text{--}} & 66.37 \\ \hdashline[0.5pt/2pt] \addlinespace[1mm] 
        0.35          & 0.42 & 0.09 &  51 (17\%) & 68.10 \\
        0.95          & 0.96 & 0.53 & 190 (65\%) & 72.43 \\
        No Unc. (all) & 1.00 & 1.00 & 294 (100\%) & 73.47 \\
        
        \bottomrule
    \end{tabular}
  }
  \vspace{-2mm}
  \caption{\textbf{Impact of $\tau$ on TT-SI.} Effect of $\tau$ on uncertain samples, efficiency, and accuracy.}
  \vspace{-3mm}
  \label{tab:uncertainty-baselines}
\end{wraptable}
.
A high $\tau$ (approaching 1) selects all samples, yielding the highest accuracy (73.47\%) but requiring updates for all 294 instances, resulting in substantial computational overhead for marginal gains.
For example, $\tau=0.95$ achieves 72.43\% accuracy with only 190 updates (35\% fewer), preserving near-optimal performance.
In contrast, a low $\tau=0.35$ minimizes false positives (FPR=0.09) but misses many errors (TPR=0.42), lowering accuracy to 68.10\%.
Thus, $\tau=0.95$ offers an effective balance, capturing most errors while avoiding redundant updates and optimizing the cost-performance trade-off, akin to human learning focused on uncertain cases (See \Cref{supplement:uncertainty-estimator} for more details on uncertainty calculations).

\begin{wrapfigure}{r}{0.4\textwidth} 
  \centering         
  \vspace{-2mm}
  \includegraphics[width=0.95\linewidth, keepaspectratio]{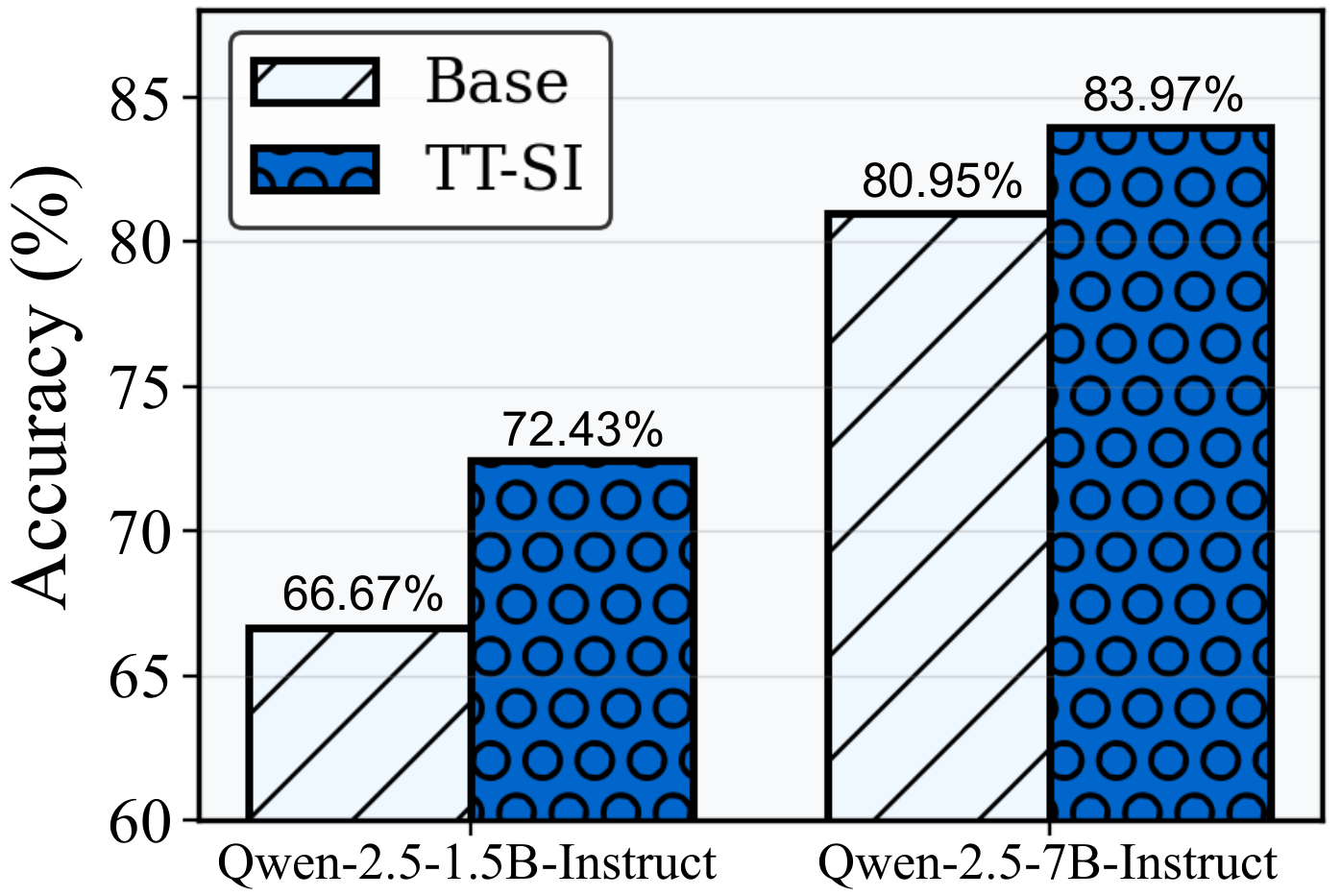} 
  \caption{\textbf{Model Scaling.} Model Scale and Architectural Generalization with \texttt{Qwen2.5-7B-Instruct}.}
  \label{fig:scale-comparison} 
  \vspace{-4mm}
\end{wrapfigure}
\noindent\textbf{Insight 7: TT-SI improves both small and large Qwen models, with larger relative gains for smaller models.} 
To assess whether TT-SI scales across architectures of different capacities, we conduct experiments with \texttt{Qwen2.5-1.5B-Instruct} and its larger counterpart \texttt{Qwen2.5-7B-Instruct} in \Cref{fig:scale-comparison}. On the smaller model, TT-SI yields a substantial +5.76\% absolute gain (66.67$\rightarrow$72.43), while on the larger model it delivers a +3.02\% gain (80.95$\rightarrow$83.97). 
These improvements indicate that TT-SI consistently enhances performance irrespective of model size, supporting its architectural generality. Interestingly, the relative boost is more pronounced for smaller models, underscoring potential of small agents as an efficiency-oriented strategy for practical deployments~\citep{belcak2025smallagents}.

\begin{wrapfigure}{r}{0.4\textwidth} 
  \centering  
  \includegraphics[width=0.95\linewidth, keepaspectratio]{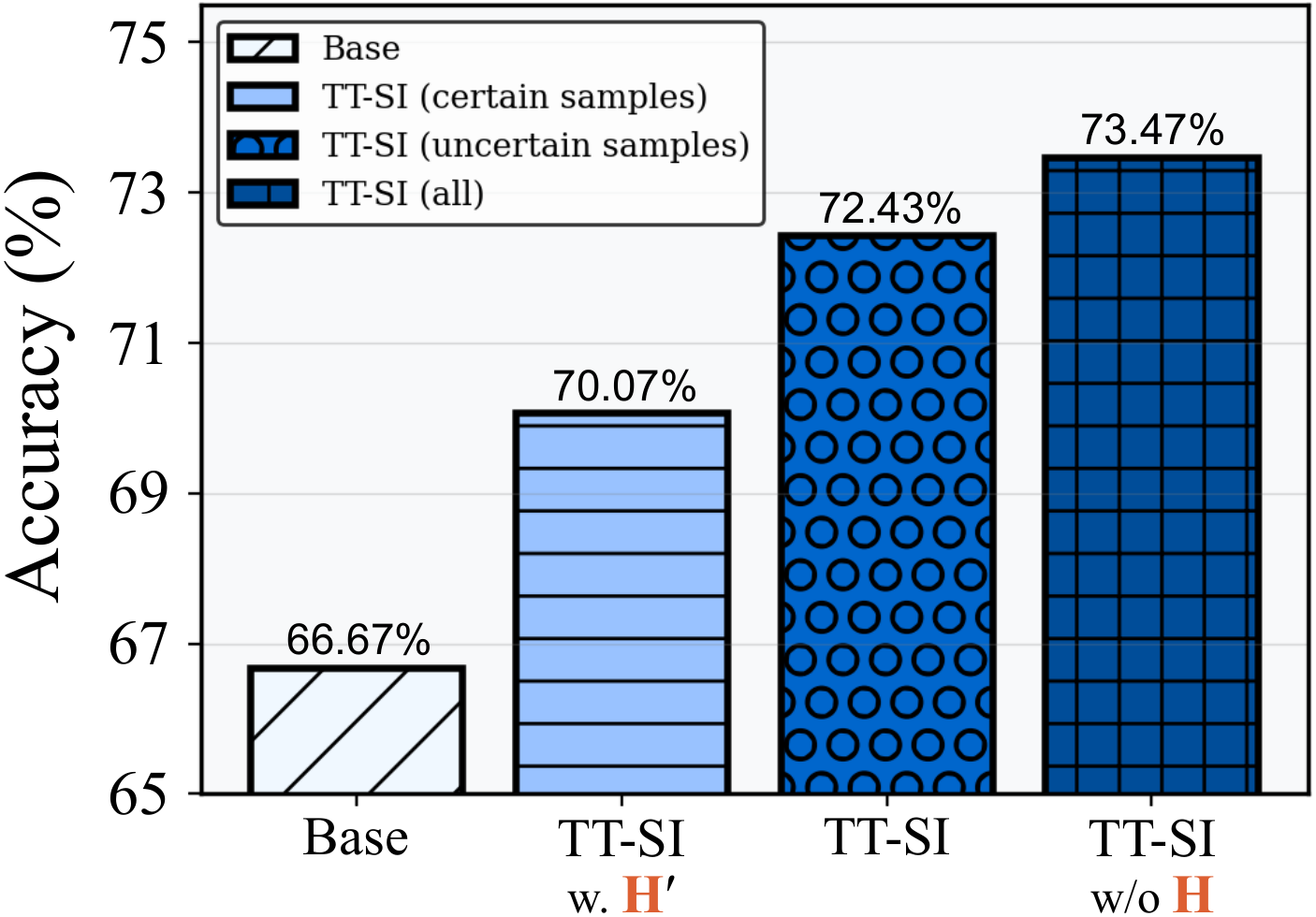} 
  \vspace{-3mm}
  \caption{\textbf{Impact of Uncertainty.} Performance of TT-SI when trained on certain samples (w. \Hunc'), uncertain samples (main), and all samples (w/o \Hunc).}
    \vspace{-4mm}
  \label{fig:h_ablation} 
\end{wrapfigure}
\noindent\textbf{Insight 8: Targeting uncertain samples is crucial for effective and efficient sharpening.}
To assess the impact of uncertain samples, we compared three variants of TT-SI on SealTool with \texttt{Qwen2.5-1.5B-Instruct}: (i) TT-SI applied only on uncertain samples detected by \Hunc\ (our main setting), (ii) TT-SI only on certain samples (w.\ \Hunc'), and (iii) TT-SI applied to all test samples regardless of uncertainty (w/o \Hunc). 
Results in \Cref{fig:h_ablation} show that focusing on uncertain samples (72.43\%) yields clear gains over using only certain samples (70.07\%),
This finding validates our core hypothesis that adaptation is most impactful on challenging instances where the model's latent knowledge is underutilized, thereby maximizing the "sharpening" effect (\Cref{eq:sharp})~\cite{huang2025selfimprovement}. While training on all samples yields a marginal gain, it incurs a prohibitive computational cost by processing 104 additional instances, undermining the core efficiency of a test-time method.

For interested readers, we further provide additional analyses in the Appendix, including more detailed \Hunc\ results (\Cref{supplement:uncertainty-estimator}), qualitative analysis of \Gsyn\ (\Cref{supplement:data_generation_details}), explore oracle effects through controlled “cheating” experiments (\Cref{supplement:cheating-exeperiment}), and discussions about runtime complexity (\Cref{supplement:time-cost}).

\vspace{-3mm}
\section{Discussions}
\vspace{-2mm}

\noindent\textbf{Conclusion\;\;} In this work, we investigate \emph{test-time self-improvement} (TT-SI) for language-based agents that (i) measures uncertainty with \uncertname\ (\Hunc) to decide whether to act directly or adapt, (ii) when uncertain, synthesizes targeted training instances with \synthname\ (\Gsyn), and (iii) performs lightweight updates via \ittname\ (\Titt). 
We demonstrate that TT-SI improve agents performance during inference, even training with one instance.
Across different benchmarks, TT-SI consistently improves test-time performance, while achieving higher accuracy and efficiency than other standard inductive learning baselines.
We further analyze the variants of TT-SI, the impact of each component, and key takeaways. 
Our results reveal the potential of TT-SI, suggesting the promise of efficient test-time learning in the development of self-evolving agents.

\noindent\textbf{Impact Statement\;\;} Our goal is not to propose a specific uncertainty metric or data generator, but rather a novel algorithm that integrates test-time learning with self-awareness, self-augmentation, and self-improvement for agentic NLP tasks. 
The design of TT-SI is modular: stronger update rules can replace SFT within \Titt, improved uncertainty quantification can plug into \Hunc, and better data generation can substitute for \Gsyn. 
We believe that, equipped with a perfect \uncertname\ (\Hunc) that renders the model self-aware of its knowledge and capabilities, a precise \synthname\ (\Gsyn) capable of generating diverse yet distributionally aligned samples from uncertain subspaces for self-augmentation, and an effective update mechanism \ittname\ (\Titt) any scenario becomes learnable in a manner akin to human learning (e.g., a student mastering challenging concepts while preparing for an exam), thereby guiding us toward the realization of a \textit{master algorithm}.

\noindent\textbf{Limitations\;\;} While TT-SI demonstrates promising results, it has limitations. First, identifying uncertain samples requires a threshold $\tau$. Although our ablations show that performance gains are consistent across different values of $\tau$ (\Cref{subsec:ablations}), the best performance is sensitive to this choice. Principled methods for learning this threshold autonomously in uncertainty calibration domain remain an open challenge~\citep{bakman2025reconsidering}. Finally, TT-SI is inherently bounded by the capacity of the base model parameters $\theta$. If the knowledge required to solve a task is absent from the pretrained model (e.g., a newly introduced medical concept), self-improvement alone cannot recover it; in such cases, external knowledge integration through retrieval or search mechanisms can be necessary.

\noindent\textbf{Future Work\;\;} TT-SI prioritizes self-improvement with test-time learning, aiming to elicit the model's optimal performance within its existing knowledge boundary. Beyond self-improvement, a key direction is enabling \textit{self-evolution} where our proposed algorithm can serve as a foundational step towards such self-evolving agents.
Another promising direction is more adaptive data generation, where the model itself determines how many synthetic examples are needed for a given uncertain case rather than relying on fixed hyperparameters~\citep{zweiger2025seal}. 
Furthermore, our current framework optimizes only the agent, not the data generator; a co-evolutionary setup like \textit{dual-learning}, where both the agent and generator adapt to each other, could further enhance performance~\citep{zhou2025scllm}. Finally, extending TT-SI to domains such as mathematics (reasoning) or medical (knowledge) presents an opportunity to explore how domain-specific uncertainty and knowledge structures interact with self-improvement~\citep{zhao2025sirius}.

\section*{Reproducibility Statement}

We truly believe transparency is essential for future and successful research. We provide our novel algorithm in \Cref{alg:uncertainty-inference}. Since our method often operates with very limited data (e.g., a single sample trainings), relatively high variance can be expected. To address this, all experiments (including baselines) are repeated five times with different random seeds, and we report averaged results with standard deviations in \Cref{sec:results}, when appropriate. All experiments are conducted on a single NVIDIA A40 GPU. For evaluation, we use publicly available and widely used benchmarks (e.g., ToolAlpaca~\citep{tang2023toolalpaca}, API-Bank~\citep{li2023api}, SealTool~\citep{wu2024sealtools}, NexusRaven~\citep{srinivasan2023nexusraven}). 
Additional implementation details, including hyperparameters and evaluation metrics, are provided in \Cref{supplement:reproductibility}.

\bibliography{iclr2026_conference}
\bibliographystyle{iclr2026_conference}

\newpage
\appendix
\section*{Appendix}
\addcontentsline{toc}{section}{Appendix} 
\startcontents
\printcontents{}{1}{\setcounter{tocdepth}{2}}
\newpage

\section{Terminology: Self-Improving and Self-Evolving Agents}
\label{supplement:self-improve-vs-evolve}

In the context of agentic systems powered by LLMs, distinguishing between \textit{self-improving} and \textit{self-evolving} agents is crucial for understanding their capabilities, limitations, and directing more clear path for future work. We try to provide some terminology with definitions, key differences, and technical insights drawn from recent literature.
\begin{itemize}
    \item \textbf{Self-Improving Agents:} 
    Self-improving agents refer to systems that autonomously enhance their own performance on specific tasks through iterative self-refinement mechanisms, without requiring any external intervention. 
    The agent iteratively generates candidate actions, evaluates them against an internal scoring signal (e.g., can be a self-reward function), and updates its subsequent steps to align with these evaluations. Thus, self-improvement in language models is achieved by reshaping their output probability distribution to preferentially weight higher-quality responses, without incorporating external knowledge beyond what is already encoded in the model parameters~\citep{huang2025selfimprovement}.
    The scope is generally task-specific, emphasizing efficiency gains within bounded domains, such as data analysis or coding, without altering the underlying system structure.
    \item \textbf{Self-Evolving Agents:} Self-evolving agents are designed for broader, continuous adaptation across dynamic environments and sequential tasks, enabling lifelong learning and generalization. These agents evolve not only parametric components (e.g., model weights) but also non-parametric elements like memory, tools, prompts, and architecture~\citep{gao2025evolvessurvey}. This allows agents to handle open-ended scenarios, such as real-world feedback loops in interactive environments.
\end{itemize}
Overall, self-improving language models focus on optimizing specific task performance by iteratively refining their output distribution toward higher-quality responses using mechanisms, without adding new information. In contrast, self-evolving language models prioritize holistic adaptation, dynamically restructuring their knowledge or architecture to enhance generalization and handle novel environments over time.

\section{Other Examples from Self-Regulated Learning}
\label{supplement:sport-example}

\noindent\textbf{Student Homework.\;\;} Our paradigm of test-time self-improvement (TT-SI) also draws inspiration from how students engage in self-learning~\citep{zimmerman2002selfregulated}. When faced with uncertainty, such as being unsure how to solve a homework problem (\textit{self-awareness}), students often seek out related examples from textbooks and online resources (\textit{self-augmentation}) to resolve their knowledge gap and build confidence in solving similar tasks~\citep{winne1998studying_as_self_regulated} (\textit{self-improvement}). This process is closely aligned with theories of metacognition and self-regulated learning, where learners actively identify their knowledge gaps and pursue targeted resources to close them \citep{nelson1990metamemory, winne1998studying_as_self_regulated, zimmerman2002selfregulated}. Unlike classical active learning in machine learning~\citep{Settles2009ActiveLL}, which deliberately queries an oracle or annotator for novel and informative examples, our method generates additional data automatically without external supervision. Moreover, while student learning often benefits from diverse perspectives and human explanations, our approach focuses on generating semantically similar but slightly varied problem instances to refine the model’s performance. This analogy highlights the natural intuition behind TT-SI while underscoring its distinct contribution as an automated, uncertainty-driven, and cost-efficient alternative to data collection.

\noindent\textbf{Sport Analogy.\;\;} Lets also consider a running back (RB) in American football honing skills for the NFL Combine, whose performance relies heavily on lower-body strength and the rate of force development for explosive acceleration. If an athlete’s primary weakness lies in short-burst speed and rapid change-of-direction, a targeted regimen emphasizing plyometric drills, resisted sprints, and eccentric–concentric coupling work can directly address this limitation. In contrast, allocating significant training time to non-specific full-body hypertrophy (e.g., frequent bench pressing or isolated arm work) not only increases recovery demands and neuromuscular fatigue but can also lead to excess non-functional muscle mass, which may reduce stride frequency and overall sprint velocity. By diagnosing the limiting factor (\textit{self-awareness}), incorporating performance-specific drills (\textit{self-augmentation}), and progressively refining execution through repeated exposure (\textit{self-improvement}), the athlete can achieve more meaningful outputs without the performance trade-offs of untargeted training.
\section{Previous Work on Large Language Models (LLMs) and Agent Fine-tuning}
\label{supplement:agent-related-work}
The de-facto approach to equip LLMs with new capabilities is to collect task-specific corpora and fine-tune on them~\citep{kumar2025llmposttraining}, with such datasets either curated through human annotation~\citep{ouyang2022rlhf} or synthesized by LLMs~\citep{wang2023selfinstruct}. Following these advancements, LLM-based agents have emerged~\citep{yao2023react}, where models interact with external tools and APIs rather than producing text alone, which require learning tool-use skills and handling structured inputs and outputs~\citep{patil2024gorilla}. Training LLMs for such agentic skills has led to the exploration of effective dataset design and tuning strategies aimed at improving generalizability~\citep{zeng2024agenttuning, mitra2024agentinstruct, chen2024agentflan}. However, this inductive approach is prone to catastrophic forgetting when transferred across different environments, requires costly data generation pipelines, and does not guarantee consistent gains over strong base models. In contrast, to the best of our knowledge, our work introduces the first application of TTT to LLM-based agents by enabling temporary parameter updates during inference and therefore avoids catastrophic forgetting and reduces dependence on large offline datasets. Furthermore, considering training efficiency, we incorporate selective data usage by using only the most informative samples for the model, rather than redundantly training on already well-understood instances. 
\section{Uncertainty Estimation Results}
\label{supplement:uncertainty-estimator}

\begin{figure}[!h]
    \centering
    \includegraphics[width=\linewidth]{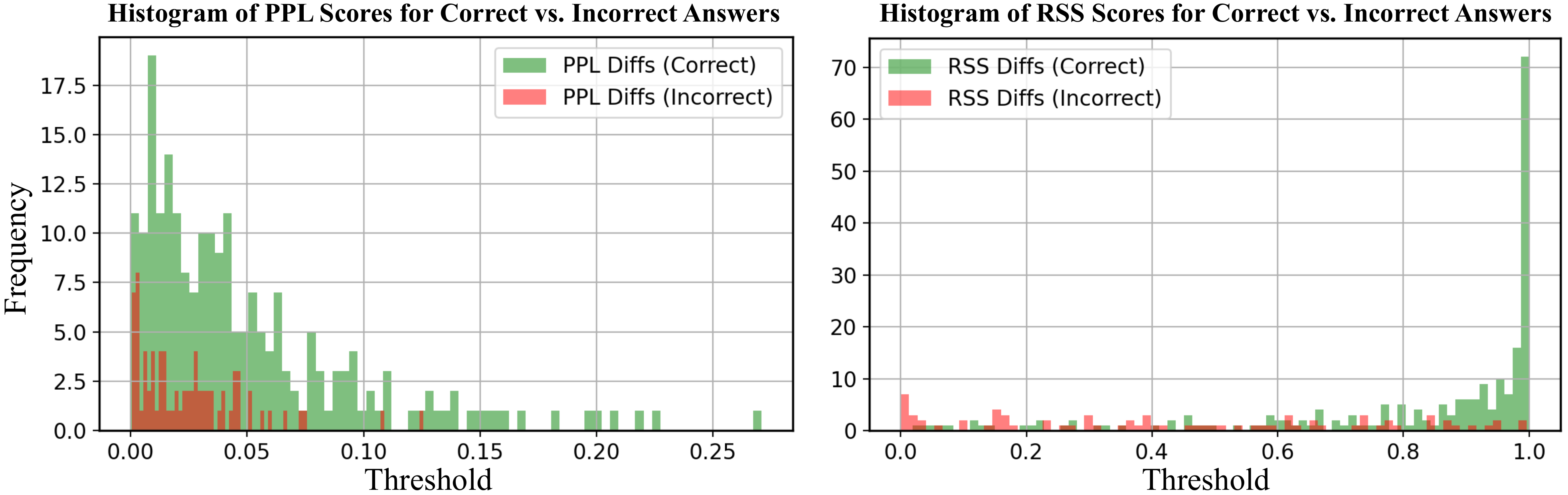}
    \caption{\textbf{Comparison of PPL and RSS (ours) as \uncertname\ (\Hunc) on SealTool.} 
Histograms show score differences between the top-1 and top-2 predictions. The 
\textcolor{darkgreen}{\textbf{green}} bars denote correct predictions, and \textcolor{red}{\textbf{red}} bars denote incorrect ones. PPL-based uncertainty (left) shows strong overlap between correct and incorrect cases, whereas our RSS-based estimator (right) yields a clearer separation, enabling more reliable uncertainty filtering.}
    \label{fig:supplement-historgram}
\end{figure}

To evaluate the effectiveness of our proposed \uncertname\ (\Hunc) in \Cref{subsec:uncertain-sample-selection}, we compare our RSS-based method defined in \Cref{eq:rss} against a standard perplexity (PPL)–based uncertainty signal using \texttt{Qwen‑2.5‑1.5B‑Instruct} on SealTool~\citep{wu2024sealtools}. We visualize the distributions of \Cref{eq:diff} as histograms in \Cref{fig:supplement-historgram}. The x-axis shows the difference between the most probable and the second most probable function call, while the y-axis denotes frequency. Green bars correspond to correctly predicted test samples, and red bars to incorrect ones.

Ideally, correct predictions (green) should cluster separately from incorrect ones (red), enabling a clear threshold $\tau$ for filtering uncertain cases. As shown in the left histogram, PPL differences for correct and incorrect answers are heavily intertwined, making it difficult to separate them with any heuristic. In contrast, our RSS-based estimator (right) exhibits a clear separation: correct predictions concentrate on the right side with larger score differences (indicating higher confidence in the top prediction), while incorrect predictions are scattered on the left side with smaller differences (reflecting model uncertainty). This separation highlights both the interpretability and effectiveness of our proposed \uncertname\ (\Hunc), especially compared to the baseline PPL-based uncertainty measure.

\begin{figure}[!h]
    \centering
    \includegraphics[width=\linewidth]{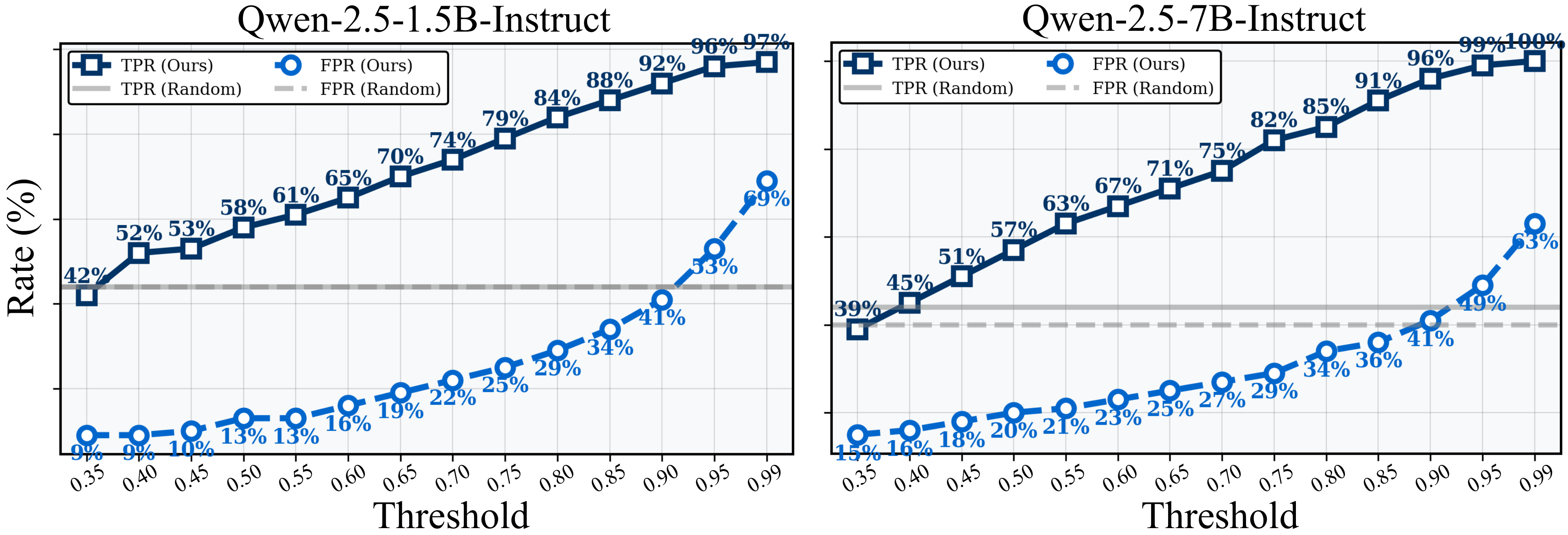}
    \caption{\textbf{Threshold ($\tau$) Experiments with \uncertname\ (\Hunc) on SealTool.} 
We investigate the effect of the softmax-difference threshold $\tau$, which controls the sensitivity of \Hunc\ in flagging uncertain cases. 
The left plot shows the true positive rate (TPR, proportion of correctly identified uncertain cases), while the right plot reports the false positive rate (FPR, proportion of certain cases incorrectly flagged as uncertain). 
As $\tau$ increases, TPR rises, but so does FPR, illustrating the trade-off between coverage and reliability.}
    \label{fig:supplement-uncertainity}
\end{figure}

We further check the performance of \uncertname\ (\Hunc) with \texttt{Qwen‑2.5‑1.5B‑Instruct} \texttt{Qwen‑2.5‑7B‑Instruct} on SealTool~\citep{wu2024sealtools} in \Cref{fig:supplement-uncertainity}. For every test input, we (i) obtain RSS confidence scores $p_n$ via~\Cref{eq:rss}, (ii) compute the softmax‑difference $u(x_i)=p^{(1)}-p^{(2)}$, and (iii) mark the prediction as \emph{uncertain} (i.e., select it for adaptation) when $u(x_i)<\tau$. We study how the choice of threshold~$\tau$ affects performance in \Cref{fig:supplement-uncertainity} by reporting true positive rate (TPR, i.e., correctly flagging the model's wrong predictions as uncertain) and false positive rate (FPR, i.e., incorrectly flagging the model's correct predictions as uncertain). As the threshold $\tau$ for the softmax-difference $u(x_i)$ increases ($0.35$ $\rightarrow$ $0.99$), the condition $u(x_i) < \tau$ for being uncertain becomes less stringent, leading to more samples being identified as uncertain and routed for downstream adaptation for both model. Raising $\tau$ monotonically increases both the TPR and FPR. Ideally, the goal is to \textbf{maximize TPR} while \textbf{keeping FPR as low as possible}. For \texttt{Qwen‑2.5‑1.5B‑Instruct}, when we increase $\tau$, TPR rises steadily from 42\% at $\tau=0.35$ to 96\% at $\tau=0.95$ and FPR remains low across all thresholds, only increasing from 9\% to 53\% as $\tau$. The overall discrimination ability of the estimator, quantified by Youden’s J statistic~\citep{ruopp2008youdenindex} (TPR $-$ FPR), reaches its highest value of 46.0\% when the threshold $\tau$ is set to 0.95. On the other hand, a similar trend is observed with \texttt{Qwen-2.5-7B-Instruct}: as $\tau$ increases, TPR rises from 39\% at $\tau=0.35$ to 99\% at $\tau=0.95$, while FPR increases from 15\% to 49\%, yielding one of the group’s highest Youden’s index values at 50.

At one of its optimal threshold, the estimator captures nearly 99\% of model errors as uncertain, while only miss classifying 49\% of correct answers. This reflects a strong balance between \textit{sensitivity} and \textit{specificity}, demonstrating the effectiveness of our (\Hunc). For all experiments, we adopt $\tau=0.95$ as the default threshold, as it consistently achieves the best trade-off between identifying the majority of erroneous outputs and minimizing unnecessary intervention on correct predictions. In \Cref{subsec:ablations}, we also analyze the impact of $\tau$ on TT-SI and find that the framework consistently improves base accuracy across all settings. Nevertheless, $\tau$ substantially influences efficiency, underscoring an inherent trade-off between accuracy and computational cost.
Moreover, $\tau$ is an hyperparameter and by tuning $\tau$, one can flexibly adjust the stringency of uncertainty filtering to match the requirements of specific downstream tasks or adaptation budgets, ensuring both effective error coverage and efficient resource allocation. Future work should establish more effective methods to automatically determine the optimal $\tau$, as this remains a central challenge in the domain of uncertainty estimation.

Finally, we compare our \Hunc\ on \texttt{Qwen-2.5-1.5B-Instruct} with several baselines on SealTool. The baselines include: Random, which labels predictions as uncertain uniformly at random; Trivial, which marks all predictions as uncertain; and Perplexity (PPL), which uses negative log-likelihood scores as an uncertainty signal based on \Cref{eq:confidence}. To evaluate, we apply each method to the ground-truth test set of SealTool and measure the resulting true positive rate (TPR), false positive rate (FPR), F1 score, and Youden’s J statistic.
\begin{wraptable}{r}{0.48\textwidth}
  \vspace{-1mm}
  \centering
  \resizebox{0.48\columnwidth}{!}{
  \begin{tabular}{lSSSS}
    \toprule
    \textbf{Method} & \textbf{TPR ($\uparrow$)} & \textbf{FPR ($\downarrow$)} & \textbf{F1 ($\uparrow$)} & \textbf{J ($\uparrow$)} \\
    \midrule
    Random       & 44.16 & 43.78 & 33.01 & 0.38 \\
    Trivial      & 100.00 & 100.00 & 41.51 & 0.00 \\
    Perplexity   & 57.14 & 38.25  & 43.14  & 18.89  \\
    \hdashline[0.5pt/2pt] \addlinespace[1mm] 
    \textbf{Ours} & 96.10 & 53.46 & 55.43 & \textbf{42.64} \\
    \bottomrule
\end{tabular}
  }
  \vspace{-2mm}
  \caption{\textbf{Comparison of Uncertainty Estimators on SealTool.} 
Our method achieves the highest balance (J) compared to Random, Trivial, and Perplexity baselines.}
\label{tab:supplement-uncertainty-baselines}
\end{wraptable}
Results show that our method achieves a TPR of 96.10\%, effectively capturing almost all misclassified samples, while maintaining a moderate FPR of 53.46\%. More importantly, our approach achieves the highest J score (42.64\%), more than double that of Perplexity (18.89\%), and also yields the best F1 score. These results quantitatively support our earlier visualization claims in \Cref{fig:supplement-historgram}, highlighting that \Hunc\ provides a strong balance between sensitivity and specificity compared to naive baselines.

\section{Data Generation Details}
\label{supplement:data_generation_details}

The implementation of \synthname\ (\Gsyn), which is triggered for each uncertain sample $x_i$, employs the agent itself for data synthesis ($\mathcal{L}_{\text{gen}}$) as \textit{self-augmentation}. For each generation instance, $\mathcal{L}_{\text{gen}}$ is provided with a carefully hand-crafted prompt $\mathcal{P}$ (See \Cref{tab:data-generation-prompt}), the uncertain input $x_i$ serving as the direct seed (critically, without its corresponding label $y_i$), and a specified number of samples $K$ to generate. The model then produces $K$ new input-output pairs, denoted as $\{(x'_{ij}, y'_{ij})\}_{j=1}^{K}$. This seed-based generation process, inspired by self-instruction methodologies~\citep{wang2023selfinstruct}, guides $\mathcal{L}_{\text{gen}}$ to produce variants that maintain the core semantic meaning and task relevance of $x_i$ while introducing controlled surface-level variations. By \textit{synthesizing data in this on-the-fly} manner for each uncertain instance, we facilitate targeted and timely model adaptation, aiming to improve performance on precisely the types of queries the model struggles with, as they are encountered.

\subsection{Experimental Results}
For each uncertain input $x_i$ detected by the procedure in \Hunc\, we synthesize exactly one new example ($K=1$) using the same LLM (i.e., \texttt{Qwen‑2.5‑1.5B‑Instruct}). The generator receives only the instruction and query of $x_i$—never the gold label—and produces both a revised input and its answer, thereby creating a temporary, query‑specific dataset $\mathcal{D}_i$ that is used immediately for inference‑time adaptation.
\begin{wrapfigure}{r}{0.5\textwidth} 
  \vspace{-\intextsep} 
  \centering         
  \includegraphics[width=0.95\linewidth, keepaspectratio]{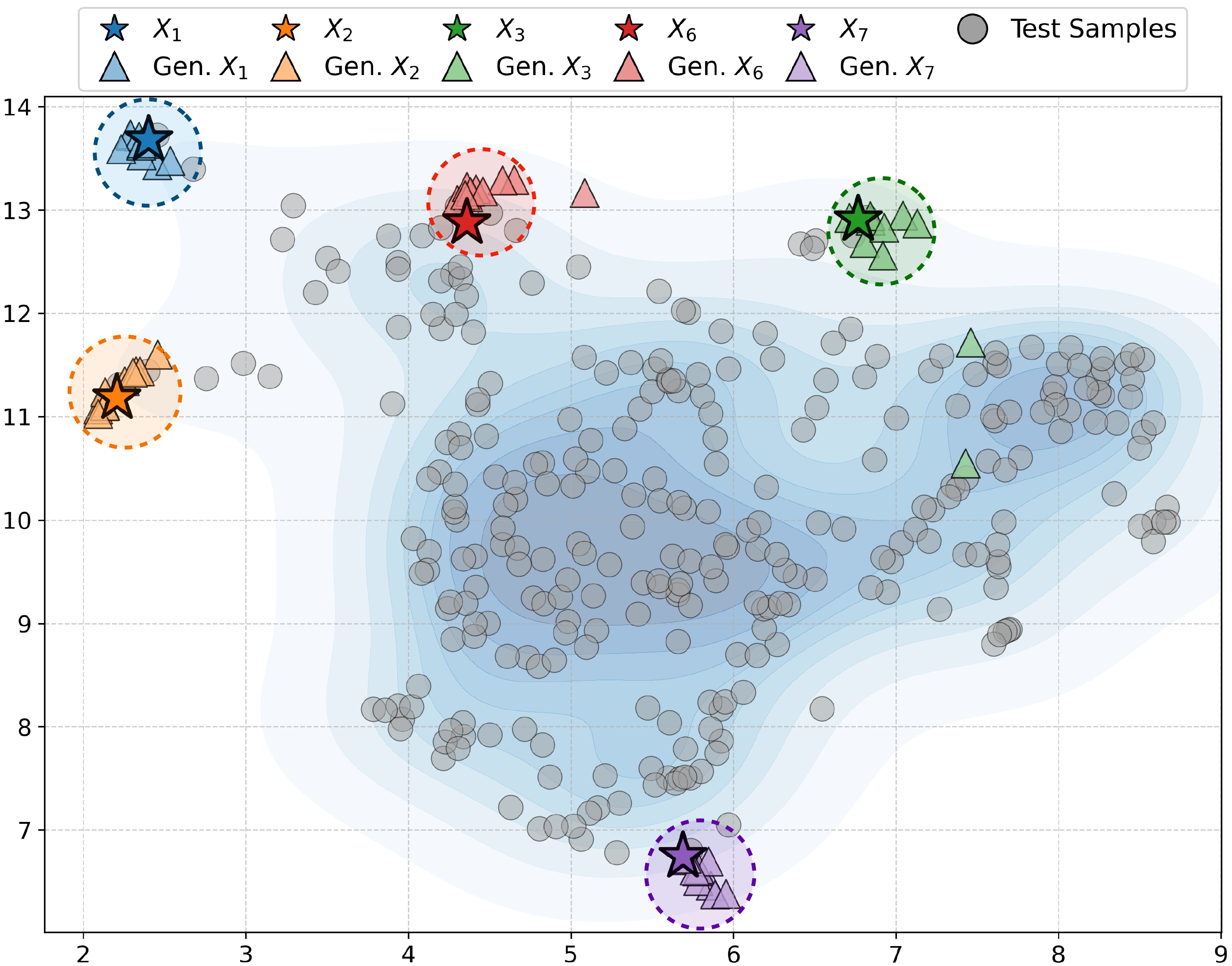} 
    \vspace{-2mm}
\caption{\textbf{Self-Generated Data Visualization.} All test samples (circles) are projected into a two-dimensional semantic space via UMAP, and shown with the density contour distributions. The star denotes the uncertain input $x_i$, and triangles indicate 10 randomly sampled, self-generated synthetic queries from uncertain sample $x_i$. Generated samples are tightly clustered and situated near $x_i$, demonstrating distributional alignment of \Gsyn.}
  \vspace{-5mm}
  \label{fig:umap-visualization} 
\end{wrapfigure}
Interpreting and understanding how our \synthname\ (\Gsyn) operates is essential for understanding the effectiveness of our generations. To this end, we embed~\citep{reimers2019sentencebert} all SealTool test samples, an uncertain example $x_i$ from this set, and ten self-generated queries for $x_i$ produced by \texttt{Qwen‑2.5‑1.5B‑Instruct} into a two-dimensional semantic space using UMAP~\citep{mcinnes2018umap}.  As visualized in \Cref{fig:umap-visualization}, the generated samples form a compact cluster in the embedding space, closely aligned with both in each other and the corresponding uncertain input. This spatial proximity suggests that our data synthesis function \Gsyn\ can yields mutually consistent and semantically faithful examples, effectively bridging the gap for adaptation to challenging queries.

\paragraph{Detailed Qualitative Analysis of Synthetic Query Generation.} To provide a more comprehensive qualitative evaluation of the quality and diversity of our self-generated synthetic queries, focus on the semantic embedding space derived from the SealTool dataset~\citep{wu2024sealtools}. We begin by encoding textual data into dense vectors. Each sample is represented as a concatenation of the system prompt, user query, and output response (or equivalent instruction-input-output triples for generated data). These are embedded using the Sentence-BERT model with \texttt{all-mpnet-base-v2}~\citep{reimers2019sentencebert}, which produces 768-dimensional vectors optimized for semantic similarity in natural language tasks. The high-dimensional embeddings are then projected into a two-dimensional latent space using Uniform Manifold Approximation and Projection (UMAP)~\citep{mcinnes2018umap}, a nonlinear dimensionality reduction algorithm that preserves both local and global topological structures more effectively than alternatives like t-SNE~\citep{vandermaaten2008tsne}. We configure UMAP with 15 neighbors to balance local clustering and global layout, and set minimum distance as 0.1 to allow moderate spread in low-density regions, facilitating the identification of outliers such as uncertain samples. 

\begin{figure*}[!h]
\begin{tcolorbox}[colback=gray!5!white,colframe=SAblue,title=\textbf{Data Generation Prompt}] 
\small You are given an instruction, input, and example sample as seed, but not labeled output. You must generate new synthetic examples that closely match the original uncertain scenario. \\

1. Create distinct variants of the seed by altering names, context, or wording but no variant may duplicate the original.\\
2. Your response format must be: \{ "instruction": "<instruction>", "input": "<input>", "output": "<output>" \} \\
3. The "output" field must be a function calling JSON object with the following structure: [\{"name": "Tool name", "parameters": \{"Parameter name": "Value",...\}\},..]  \\

\textbf{\#\#\# Number of Examples} \\
Generate <number> examples. \\
 \\
\textbf{\#\#\# Seed Example} \\
<seed\_example> \\
 \\
\textbf{\#\#\# Generated Examples} \\
Your Response:
\end{tcolorbox}

\vspace{-0.25cm}
\caption{Data Generation Prompt for uncertain samples with LLM.}
\vspace{-6mm}
\label{tab:data-generation-prompt}
\end{figure*}

\section{Cheating Experiments and TT-SI Comparison}
\label{supplement:cheating-exeperiment}

\begin{figure}
    \centering
    \includegraphics[width=0.85\linewidth]{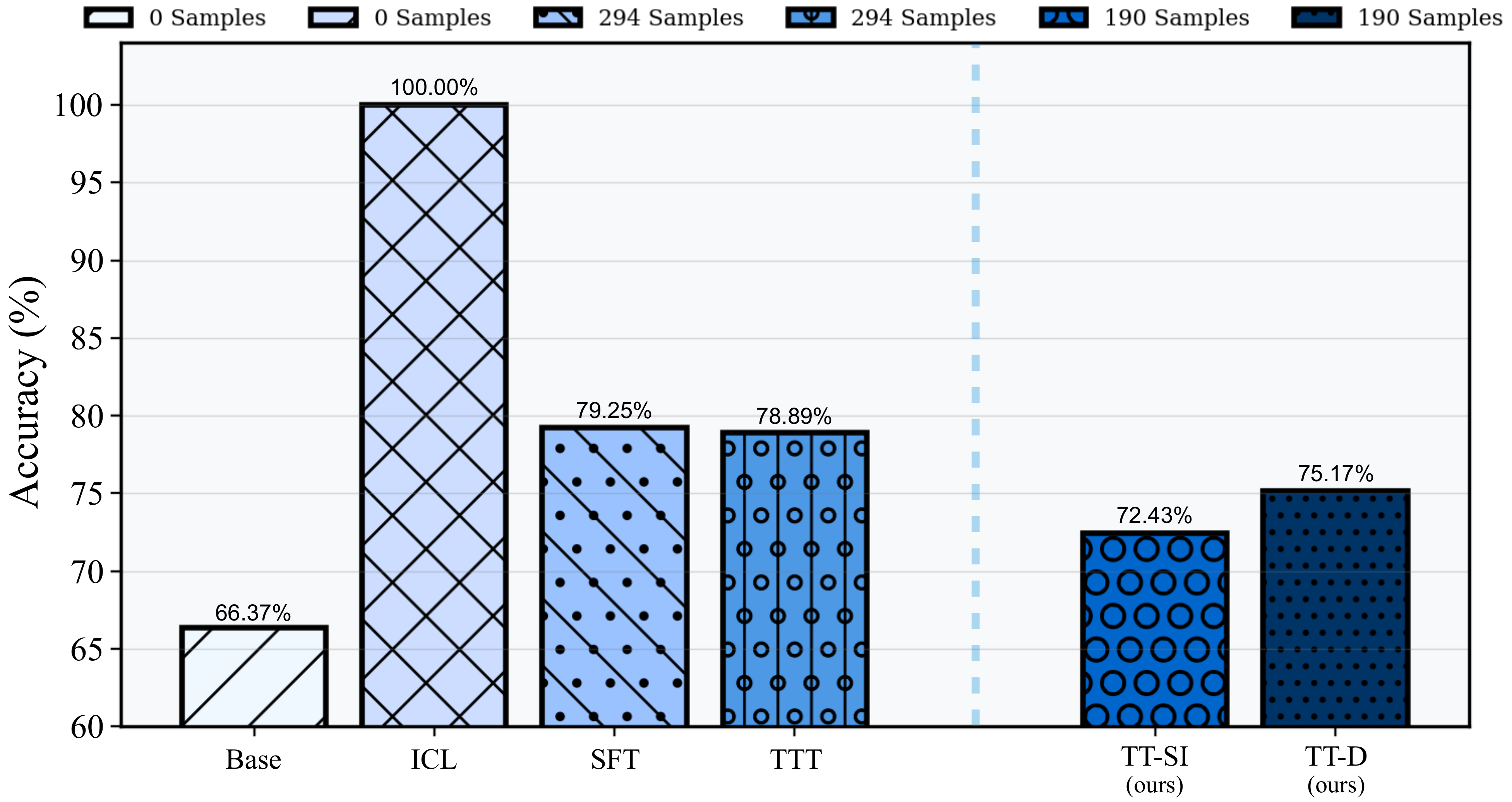}
    \caption{\textbf{Cheating Experiment on SealTool.} Comparison of baseline methods—Base, in-context learning (ICL), supervised fine-tuning (SFT), and test-time training (TTT)—when explicitly trained on the test set (left four bars) using \texttt{Qwen-2.5-1.5B-Instruct}. We also report actual (non-cheating) scores for our TT-SI algorithm and its TT-D variant (right two bars).}
    \label{fig:supplement-cheat}
\end{figure}

To better contextualize our proposed method, we conducted a \emph{cheating experiment} in which baseline models were explicitly trained on the test set. This unrealistic setting serves as an upper bound on performance for common adaptation strategies, including in-context learning (ICL), supervised fine-tuning (SFT), and test-time training (TTT). Using \texttt{Qwen-2.5-1.5B-Instruct}, we report results on the SealTool benchmark in \Cref{fig:supplement-cheat}, where the left four bars show the cheating baselines. For comparison, we include our proposed TT-SI framework and its TT-D variant (right bars), which were not trained on the test set but instead evaluated under their standard configuration. 

When comparing cheating TTT (78.89\%) with our TT-SI (72.43\%), the scores are remarkably close, suggesting that providing highly similar samples during test-time training (rather than exact ground truth answers) is sufficient to shift the model toward the uncertain sample distribution. This process increases confidence for samples previously overlooked by the base model's parameters, leading to improved performance after temporary updates. If the gap between cheating TTT and TT-SI were larger, it would indicate that exact ground truth answers are critical and that better data generation methods beyond self-improvement are needed. Interestingly, TT-SI achieves scores comparable to SFT trained on the test set. More surprisingly, neither update-based method (SFT, TTT) reaches 100\% accuracy after one epoch of training on actual samples, suggesting issues with the update rules themselves. Also, it is important to mention that SFT reaches 96.60\% after 10 epochs of training on this 294 sample test set. In contrast, ICL achieves 100\% accuracy directly when these actual test samples are added to the prompt during inference.

\section{Implementation Details of TT-SI}
\label{supplement:reproductibility}

We firmly believe that transparency and detailed reporting are essential for both understand the approach in-depth and advancing future research. Accordingly, we do our best to provide complete descriptions of each component of the proposed TT-SI framework: \uncertname\ (\Hunc) (\Cref{subsec:uncertain-sample-selection}), \synthname\ (\Gsyn) (\Cref{subsec:synthesize-similar-samples}), and \ittname\ (\Titt) (\Cref{subsec:inference_time_finetuning}). In all steps, we use \texttt{Qwen2.5-1.5B-Instruct}\footnote{\url{https://huggingface.co/Qwen/Qwen2.5-1.5B-Instruct}} from HuggingFace checkpoints, running on a single NVIDIA A40 GPU.

\subsection{Uncertainty Estimation}
\label{supplement:reproductibility-step1}
For implementing \Hunc, we use the HuggingFace Transformers library~\citep{wolf2019huggingface}, as it provides straightforward access to token logits and confidence estimates through the \texttt{AutoModelForCausalLM} class, unlike vLLM. We do not apply temperature scaling at this step. To estimate uncertainty, we directly input the test sample instruction as a query and merge all available function names extracted via regex operations. The confidence scores for each candidate function are computed using \Cref{eq:confidence} and normalized with RSS as formulated in \Cref{eq:rss}. On SealTool, labeling a sample as uncertain requires on average $0.87$ seconds.

\subsection{Data Generation}
Once an uncertain sample is identified with \Hunc, we generate $K$ similar samples using \Gsyn. This is done with the prompt shown in \Cref{tab:data-generation-prompt}, where the model is asked to create slight variations of the sample (but not the exact same sample) along with corresponding labels. The uncertain sample is inserted into the prompt as a seed (\texttt{<}seed\texttt{>})~\citep{wang2023selfinstruct}, and $K$ is set as a hyperparameter (\texttt{<}number\texttt{>}) by replacing special tokens. We then extract the generated samples. We study two variants: TT-SI and TT-D. In TT-SI, the same \texttt{Qwen2.5-1.5B-Instruct} model generates its own synthetic samples, while in TT-D, sample generation is performed by \texttt{GPT-5-mini}\footnote{\url{https://platform.openai.com/docs/models/gpt-5-mini}}. For TT-SI, we use vLLM with temperature $0.7$ and maximum length set to $32768$. Because of the model’s small scale, sometimes parsing errors occur, where the model may omit some JSON strings. We allow up to 5 retries; in practice, errors are usually resolved within the second or third attempt. For \texttt{gpt-5-mini}, we use the standard OpenAI API without temperature adjustments or additional decoding strategies. The computational cost of using \texttt{gpt-5-mini} API is negligible, effectively zero for a single experiment. On average, generating one sample with TT-SI takes approximately $3.45$ seconds. A qualitative analysis of the data generation process is provided in \Cref{supplement:data_generation_details}.

\subsection{Training}
After obtaining the $K$ generated samples from \Gsyn, we directly perform test-time fine-tuning with \Titt. For training, we use LLaMA-Factory~\citep{zheng2024llamafactory}, chosen for its optimized implementation and user-friendly CLI. All fine-tuning is conducted with Parameter-Efficient Fine-Tuning (PEFT) via LoRA~\citep{hu2022lora}. We set the LoRA parameters to $rank=8$ and $\alpha=16$, applied to all linear layers. Training runs for 5 epochs with a fixed learning rate of $1.0\times10^{-4}$, a warm-up ratio of $0.03$, and batch size of 1. Despite the short training, we use a cosine scheduler by default. Otherwise, we keep the default configuration parameters of LLaMA-Factory and HuggingFace without further modifications. For inference on the fine-tuned models, we adopt the same vLLM decoding settings as in data generation with temperate is set to $0.7$. All trainings follow the Alpaca-style data format~\citep{alpaca}, where the instruction and input fields are zero-padded, and the loss is computed only on the output field. Training a single sample with LLaMA-Factory takes on average $2.05$ seconds. On the other hand, our reported timings exclude software initializations,  I/O operation overheads from checkpoint loading, merging, and saving, as these are highly implementation/tool-dependent, which is discussed in \Cref{supplement:time-cost}.

\subsection{Evaluation}
We evaluate our method on three established agent benchmarks: NexusRaven~\citep{srinivasan2023nexusraven}, SealTool~\citep{wu2024sealtools}, API-Bank~\citep{li2023api}, and ToolAlpaca~\citep{tang2023toolalpaca}.
\textbf{NexusRaven} focuses on realistic software operation tasks, particularly in domains such as cybersecurity and enterprise applications. It is designed to test high-fidelity function execution in business scenarios, featuring long and diverse tool invocations across 65 distinct APIs with a total of 318 samples (see \Cref{tab:nexusraven} for an example).
\textbf{SealTool} is one of the most extensive and recent benchmarks, comprising 4,076 APIs spanning diverse domains. Its latest version is designed to minimize potential data leakage, making it a robust benchmark for tool-use evaluation. In our experiments, we use the curated test set of 294 samples (see \Cref{tab:sealtool} for an example).
\textbf{API-Bank} contains 314 multi-turn conversations with 753 distinct API calls. It evaluates an LLM’s ability to select appropriate functions and arguments in realistic dialogue settings. Following prior work, we focus on 316 samples from Levels 1 and 2, which balance task complexity and data availability (see \Cref{tab:apibank} for an example).
\textbf{ToolAlpaca} employs a synthetic data generation framework, featuring 3,938 tool-use instances in 50 categories, designed to assess generalized tool-use capabilities across diverse APIs (see \Cref{tab:toolalpaca} for an example).

Following prior work~\citep{lin2025hammer}, we use 318, 294, 361, and 103 test samples for each benchmark, respectively, consistent with previous studies, except for slight modifications on evaluation metrics to ensure a more accurate and reliable evaluation setup.
Across all benchmarks, we evaluate whether models produce correct function names, arguments, and their corresponding values/types. A key challenge involves string arguments, where models often produce superficial variations of gold-standard values—differing in case, tense, or plurality (e.g., "fatigued" vs. "fatigue" or "fatigous"). We argue these discrepancies reflect artifacts of current benchmarks rather than genuine errors, yet they are difficult to evaluate fairly: exact-match metrics are overly strict, while LLM-based judges introduce unreliability. We therefore adopt a soft-matching metric that ignores case and minor morphological variations. This adjustment changes performance by only 2–3\%, but provides a more accurate estimate of functional correctness. Thus, our evaluation framework prioritizes semantic equivalence over superficial string matching, better reflecting real-world tool-calling capability.

\section{Additional Run-time Overhead and Resource Usage Analysis}
\label{supplement:time-cost}

We analyze the latency of each algorithmic step of TT-SI on the SealTool dataset, excluding model merging and other I/O overheads, which we discuss separately below. For consistency, we employ HuggingFace~\citep{wolf2019huggingface} for confidence estimation with \Hunc, vLLM~\citep{kwon2023vllm} for data synthesis with \Gsyn\ and inference, and LLaMA-Factory~\citep{zheng2024llamafactory} for trainings with \Titt. 

Table~\ref{tab:latency} reports total latency, per-sample averages, and variation statistics. On average, \Hunc\ requires $0.87s$ per sample to estimate uncertainty. For uncertain inputs, \Gsyn\ synthesizes additional variants in $3.45s$, which are subsequently used for \Titt\ training updates that take $2.05s$ each. Finally, inference via vLLM adds only $0.89s$ per sample. These steps together amount to an average of $\sim7.3s$ per uncertain sample, while non-uncertain samples require only $0.87s$; amounting to $\sim$36 minutes for 190 updates and 104 direct inferences.  In contrast, SFT requires $7,966.6s$ ($\sim$2h12m) to train on SealTool’s 13K-sample split. Despite training on $\sim$68$\times$ fewer samples, TT-SI delivers a 3.7$\times$ wall-clock speed-up. 

\begin{table}[!h]
\centering
\small
\begin{tabular}{lccccc}
\toprule
\textbf{Step} & \textbf{Total (s)} & \textbf{Avg (s)} & \textbf{Std (s)} & \textbf{Min (s)} & \textbf{Max (s)} \\
\midrule
Uncertainty (\Hunc) & 254.37 & 0.87 & 0.13 & 0.53 & 1.20 \\
Data Generation (\Gsyn) & 644.37 & 3.45 & 1.23 & 2.04 & 9.63 \\
Training (\Titt) & 389.87 & 2.05 & 0.42 & 1.67 & 3.49 \\
Inference & 260.54 & 0.89 & 0.14 & 0.49 & 1.34 \\
\hdashline[0.5pt/2pt] \addlinespace[1mm]
\multicolumn{6}{c}{\textbf{Total per-sample: 7.26s (uncertain) \quad | \quad 1.76s (certain)}} \\
\bottomrule
\end{tabular}
\caption{\textbf{Latency Analysis.} Breakdown of TT-SI step-wise latency on SealTool, with merge and file I/O overhead excluded. The bottom row reports the end-to-end average latency: $7.26$s for uncertain samples and $1.76$s for certain samples. Under $\tau=0.95$, TT-SI processes 190 uncertain and 194 certain samples on SealTool.}
\label{tab:latency}
\end{table}

However, we note that most of the additional latency stems from model merging, file-saving operations after training, and vLLM model loading. While our main algorithmic steps: \uncertname\ (\Hunc), \synthname\ (\Gsyn), and \ittname\ (\Titt) introduce only minimal computational overhead as discussed above, I/O operations can substantially increase end-to-end latency. Since efficient file handling lies outside the scope of our main contribution, we do not focus on these issues in our work. For these reason, we exclude such I/O overheads from the reported clock-time analysis. On the other hand, third-party libraries offer options to directly use merged weights without writing them to disk, and more efficient configurations can be implemented to manage these steps. Thus, we recommend that future industrial applications prioritize more optimized and scalable I/O strategies.
\section{Use of LLMs}
In this work, LLMs were used in this work for three purposes: (i) as base models under study for test-time training, (ii) as baselines for empirical comparison, and (iii) for minor assistance in refining the readability of this manuscript. Both open-source (e.g., Qwen~\citep{yang2025qwen3}) and closed-source models (e.g., GPT-5-mini\footnote{\url{https://platform.openai.com/docs/models/gpt-5-mini}}) were employed for training and data-generation. 
The prompt used for improving writing quality was similar to \textit{"Please make more clear sentence, making sure to remove any grammatical mistakes."}
Importantly, all scientific ideas, methods, experiments, and conclusions originate from the authors. When LLMs were used for language refinement, outputs were carefully reviewed to prevent the introduction of hallucinated or incorrect content, ensuring that all arguments, findings, and perspectives are solely those of the authors.

\clearpage

\begin{figure*}[!h]
\begin{tcolorbox}[
    colback=gray!5!white,
    colframe=SAblue,
    title={\textbf{NexusRaven Test Sample Example (ID: 317)}},
    fontupper=\scriptsize, 
    boxrule=0.5mm,
    width=\textwidth 
    ]
{\scriptsize 
You are an advanced assistant capable of using tools to help the user. You may call one or more functions to assist with the user query.\\
For any user request that requires a function, respond by returning a function call inside \textbf{<tool\_call>}...\textbf{</tool\_call>} XML tags, with a JSON object specifying the "name" of the function and the "arguments".\\
\\
\textbf{Task Instruction}\\
In order to complete the user's request, you need to select one or more appropriate tools from the following tools and fill in the correct values for the tool parameters. Your specific tasks are:\\
1. Make one or more function/tool calls to meet the request based on the question.\\
2. If none of the functions can be used, point it out as an empty list and refuse to answer.\\
3. If the given question lacks the parameters required by the function, also point it out.\\
\\
\textbf{Output Format}\\
For each function call, return a JSON object with function name and arguments within \textbf{<tool\_call></tool\_call>} XML tags:\\
\textbf{<tool\_call>}[ \{"name": "<function-name>", "arguments": \{"arg1": "value1", "arg2": "value2", ...\} }, ...] \textbf{</tool\_call>}\\
If no function call is needed, please directly output an empty list `\textbf{[]}' as \textbf{<tool\_call>[]</tool\_call>}.\\
\\
\textbf{Available Tools}:\\
In your response, you can use the following tools:\\
\textbf{<tools>}\\
1. Name: verifyUSAddress\\
Description: Verify a given US address to ensure it meets USPS standards and is deliverable.\\
Parameters: \{`addressLine1': \{`type': `str', `description': `The primary address line, including street number and name.', `required': True\}, `addressLine2': \{`type': `str', `description': `The secondary address line, such as apartment or suite number.', `required': True\}, `city': \{`type': `str', `description': `The city of the address.', `required': True\}, `state': \{`type': `str', `description': `The state or territory of the address.', `required': True\}, `zipCode': \{`type': `str', `description': `The 5-digit ZIP code of the address.', `required': True\}\}\\
2. Name: standardizeUSAddress\\
Description: Standardize a given US address to create consistency and accuracy in addressing.\\
Parameters: \{`addressLine1': \{`type': `str', `description': `The primary address line, including street number and name.', `required': True\}, `addressLine2': \{`type': `str', `description': `The secondary address line, such as apartment or suite number.', `required': True\}, `city': \{`type': `str', `description': `The city of the address.', `required': True\}, `state': \{`type': `str', `description': `The state or territory of the address.', `required': True\}, `zipCode': \{`type': `str', `description': `The 5-digit ZIP code of the address.', `required': True\}\}\\
\textbf{</tools>}\\
\\
\textbf{Question}\\
User: I'm organizing a mailing list for my business, and I want to make sure all the addresses are standardized. Can you help me standardize this address? 456 Street, Suite 7891, Los Angeles, CA, 90011.\\
\\
\textbf{Your Response}: \textbf{<tool\_call>}[ \{"name": "standardizeUSAddress", "arguments": \{"addressLine1": "456 Street", "addressLine2": "Suite 7891", "city": "Los Angeles", "state": "CA", "zipCode": "90011"\} \} ] \textbf{</tool\_call> } 
\end{tcolorbox}
\caption{Sample example from NexusRaven test data.}
\label{tab:nexusraven}
\end{figure*}

\begin{figure*}[!h]
\begin{tcolorbox}[
    colback=gray!5!white,
    colframe=SAblue,
    title={\textbf{SealTool Test Sample Example (ID: 4)}},
    fontupper=\scriptsize, 
    boxrule=0.5mm,
    width=\textwidth 
    ]
{\scriptsize 
You are an advanced assistant capable of using tools to help the user. You are given a conversation between a user and an assistant, together with the available tools.\\
You may call one or more functions to assist with the user query.\\
You will be provided with a set of Available Functions inside \textbf{<tools>...</tools>} tags.\\
For any user request that requires a function, respond by returning a function call inside \textbf{<tool\_call>...</tool\_call>} XML tags, with a JSON object specifying the "name" of the function and the "arguments".\\
\\
\textbf{Task}\\
1. Think and recall relevant context, analyze the current user goal.\\
2. Refer to the previous dialogue records in the conversations, including the user's queries.\\
3. Decide on which tool to use from \textbf{Available Tools} and specify the tool name.\\
4. At the end, you need to output the JSON object of the function call inside the \textbf{<tool\_call>} and \textbf{</tool\_call>} tags.\\
5. Output format of the function calls must be EXACTLY like in the \textbf{Output Format} section, the function calls must be a list of JSON objects, each object must have a "name" key and an "arguments" key.\\
6. This year is 2023.\\
\\
\textbf{Output Format}\\
For each function call, return a JSON object with function name and arguments within \textbf{<tool\_call></tool\_call>} XML tags:\\
\textbf{<tool\_call>}[ \{"name": "<function-name>", "arguments": \{"arg1": "value1", "arg2": "value2", ...\} }, ...] \textbf{</tool\_call>}\\
\\
\textbf{Available Tools}\\
\textbf{<tools>}\\
1. Name: analyzeSample\\
Description: Analyze a given sample using analytical chemistry techniques\\
Field: Chemistry/Analytical chemistry\\
Parameters: \{`sample': \{`type': `str', `description': `The sample to be analyzed'\}, `method': \{`type': `str', `description': `The analytical method to be used for analysis (e.g., chromatography, spectroscopy)'\}, `instrument': \{`type': `str', `description': `The instrument or equipment to be used for analysis (e.g., gas chromatograph, mass spectrometer)'\}, `conditions': \{`type': `str', `description': `Any specific conditions required for the analysis (e.g., temperature, pressure)'\}\}\\
Required: [sample, method]\\
Responses: \{`results': \{`type': `str', `description': `The analysis results containing information about the sample'\}\}\\
2. Name: analyzeEvidence\\
Description: Analyze the chemical evidence collected from a crime scene\\
Field: Chemical Engineering/Forensic engineering\\
Parameters: \{`evidence\_type': \{`type': `str', `description': `The type of evidence to be analyzed (e.g., DNA, fingerprints, blood, fibers)'\}, `method': \{`type': `str', `description': `The method or technique to be used for analysis (e.g., spectroscopy, chromatography, microscopy)'\}, `sample': \{`type': `str', `description': `The sample or specimen to be analyzed (e.g., crime scene swab, hair strand, fabric sample)'\}\}\\
Required: [evidence\_type, method, sample]\\
Responses: \{`analysis\_results': \{`type': `str', `description': `The results of the chemical analysis of the evidence'\}, `conclusion': \{`type': `str', `description': `The conclusion drawn from the analysis'\}\}\\
3. Name: getSampleSize\\
Description: Retrieve the sample size of a mixed methods research study\\
Field: Research/Mixed Methods Research\\
Parameters: \{`study\_id': \{`type': `str', `description': `The unique identifier of the research study'\}\}\\
Required: [study\_id]\\
Responses: \{`sample\_size': \{`type': `int', `description': `The sample size of the research study'\}\}\\
4. Name: getFabricComposition\\
Description: Retrieve fabric composition information for a specific clothing item\\
Field: Fashion/Fashion Technology\\
Parameters: \{`clothing\_item': \{`type': `str', `description': `The type of clothing item for which you want fabric composition (e.g., t-shirt, jeans, dress)'\}, `brand': \{`type': `str', `description': `The brand of the clothing item (e.g., Nike, Zara, Gucci)'\}\}\\
Required: [clothing\_item]\\
Responses: \{`composition': \{`type': `str', `description': `The fabric composition of the specified clothing item'\}, `brand': \{`type': `str', `description': `The brand of the clothing item'\}\}\\
5. Name: evaluateDataBias\\
Description: Evaluate data bias in a dataset\\
Field: Data Analysis/Data Ethics\\
Parameters: \{`dataset': \{`type': `str', `description': `The dataset to evaluate for bias (e.g., hiring records, loan applications)'\}, `protected\_attributes': \{`type': `str', `description': `The protected attributes to consider for bias assessment (e.g., gender, race)'\}, `measures': \{`type': `str', `description': `The bias assessment measures to be used (e.g., disparate impact, statistical parity index)'\}, `reference\_group': \{`type': `str', `description': `The reference group to compare with for bias assessment'\}\}\\
Required: [dataset, protected\_attributes]\\
Responses: \{`bias\_score': \{`type': `float', `description': `The overall bias score of the dataset'\}, `protected\_attributes\_bias': \{`type': `str', `description': `Detailed bias assessment for each protected attribute'\}\}\\
\textbf{</tools>}\\
\\
\textbf{Input}\\
User: Provide the statistics for the Real Madrid team.\\
\\
\textbf{Your Response}: \textbf{<tool\_call>}[ \{"name": "getTeamStats", "arguments": \{"team": "Real Madrid"\} \} ] textbf{</tool\_call> }
\end{tcolorbox}
\caption{Sample example from SealTool test data.}
\label{tab:sealtool}
\end{figure*}

\begin{figure*}[!h]
\begin{tcolorbox}[
    colback=gray!5!white,
    colframe=SAblue,
    title={\textbf{API-Bank Test Sample Example (ID: 0)}},
    fontupper=\scriptsize, 
    boxrule=0.5mm,
    width=\textwidth 
    ]
{\scriptsize 
You are an advanced assistant capable of using tools to help the user. You are given a conversation between a user and an assistant, together with the available tools.\\
You may call one or more functions to assist with the user query.\\
For any user request that requires a function, respond by returning a function call inside \\textbf{texttt}{<tool\_call>...</tool\_call>} XML tags, with a JSON object specifying the "name" of the function and the "arguments".\\
\\
\textbf{Task}\\
1. Think and recall relevant context, analyze the current user goal.\\
2. Refer to the previous dialogue records in the conversations, including the user's queries.\\
3. Decide on which tool to use from \textbf{Available Tools} and specify the tool name.\\
4. At the end, you need to output the JSON object of the function call inside the \textbf{<tool\_call>} and \textbf{</tool\_call>} tags.\\
5. Output format of the function calls must be EXACTLY like in the \textbf{Output Format} section, the function calls must be a list of JSON objects, each object must have a "name" key and an "arguments" key.\\
6. This year is 2023.\\
\\
\textbf{Output Format}\\
For each function call, return a JSON object with function name and arguments within \textbf{<tool\_call></tool\_call>} XML tags:\\
\textbf{<tool\_call>}[ \{"name": "<function-name>", "arguments": \{"arg1": "value1", "arg2": "value2", ...\} }, ...] \textbf{</tool\_call>} \\
\\
\textbf{Available Tools}:\\
In your response, you can use the following tools:\\
\textbf{<tools>}\\
1. Name: QueryHealthData\\
Description: This API queries the recorded health data in database of a given user and time span.\\
Parameters: \{`user\_id': \{`type': `str', `description': `The user id of the given user. Cases are ignored.'\}, `start\_time': \{`type': `str', `description': `The start time of the time span. Format: \%Y-\%m-\%d \%H:\%M:\%S'\}, `end\_time': \{`type': `str', `description': `The end time of the time span. Format: \%Y-\%m-\%d \%H:\%M:\%S'\}\}\\
2. Name: CancelRegistration\\
Description: This API cancels the registration of a patient given appointment ID.\\
Parameters: \{`appointment\_id': \{`type': `str', `description': `The ID of appointment.'\}\}\\
3. Name: ModifyRegistration\\
Description: This API modifies the registration of a patient given appointment ID.\\
Parameters: \{`appointment\_id': \{`type': `str', `description': `The ID of appointment.'\}, `new\_appointment\_date': \{`type': `str', `description': `The new appointment date. Format: \%Y-\%m-\%d.'\}, `new\_appointment\_doctor': \{`type': `str', `description': `The new appointment doctor.'\}\}\}\\
\textbf{</tools>}\\
\\
\textbf{Conversation}\\
User: Can you please modify my appointment scheduled for March 25th with Dr. Kim to March 26th with Dr. Lee?\\
Assistant: Sure, I can help you with that. Please provide me with the appointment ID and the new appointment date and doctor's name.\\
User: The appointment ID is 34567890 and the new date is March 26th with Dr. Lee.\\
Assistant: Alright. I'll modify your appointment now.\\
User: Based on our conversation above, please only make one tool call to solve my need.\\
\\
\textbf{Output}: \textbf{[<tool\_call>}[\{"name": "ModifyRegistration", "arguments": \{"appointment\_id": "34567890", "new\_appointment\_date": "2023-03-26", "new\_appointment\_doctor": "Dr. Lee"\}\}]\textbf{</tool\_call>] }
\end{tcolorbox}
\caption{Sample example from API-Bank test data.}
\label{tab:apibank}
\end{figure*}

\begin{figure*}[!h]
\begin{tcolorbox}[
    colback=gray!5!white,
    colframe=SAblue,
    title={\textbf{ToolAlpaca Test Sample Example (ID: 35)}},
    fontupper=\scriptsize, 
    boxrule=0.5mm,
    width=\textwidth 
    ]
{\scriptsize 
You are an advanced assistant capable of using tools to help the user. \\
You may call one or more functions to assist with the user query.\\
You will be provided with a set of Available Functions inside \textbf{<tools>}...\textbf{</tools>} tags.\\
For any user request that requires a function, respond by returning a function call inside \textbf{<tool\_call>}...\textbf{</tool\_call>} XML tags, with a JSON object specifying the "name" of the function and the "arguments".\\
\\
\textbf{Task Instruction}\\
In order to complete the user's request, you need to select one or more appropriate tools from the following tools and fill in the correct values \\
for the tool parameters. Your specific tasks are:\\
    1. Make one or more function/tool calls to meet the request based on the question.\\
    2. If none of the function can be used, point it outas empty list and refuse to answer.\\
    3. If the given question lacks the parameters required by the function, also point it out.\\
\\
\textbf{Output Format}\\
For each function call, return a JSON object with function name and arguments within \textbf{<tool\_call>}\textbf{</tool\_call>} XML tags:\\
\textbf{<tool\_call>}[ \{"name": "<function-name>", "arguments": \{"arg1": "value1", "arg2": "value2", ...\} }, ...]</tool\_call>\\
If no function call is needed, please directly output an empty list '[]' as \textbf{<tool\_call>}[]\textbf{</tool\_call>}\\
\\
\textbf{Available Tools}:\\
In your response, you can use the following tools:\\
\textbf{<tools>}
1. Name: airports\_get\\
Description: Get an airport by its ICAO or FAA identifier\\
Parameters: \{'apt': \{'type': 'string', 'description': 'FAA or ICAO facility identifier (KAVL or AVL). Separate multiple entries with a comma. Required is true.', 'required': True\}\}\\
2. Name: charts\_get\\
Description: Get charts for a specified airport\\
Parameters: \{'apt': \{'type': 'string', 'description': 'FAA or ICAO airport identifier (KAVL or AVL). Separate multiple entries with a comma. Required is true.', 'required': True\}, 'group': \{'type': 'integer', 'description': 'Optional grouping of the charts. 1 -> General, Departures, Arrivals, Approaches; 2 -> Airport Diagram only; 3 -> General only; 4 -> Departures only; 5 -> Arrivals only; 6 -> Approaches only; 7 -> Everything but General.'\}\}\\
3. Name: charts\_changes\_get\\
Description: Get chart changes by airport or chart name\\
Parameters: \{'apt': \{'type': 'string', 'description': 'FAA or ICAO airport identifier (KAVL or AVL). Required is true.', 'required': True\}, 'chart\_name': \{'type': 'string', 'description': 'Partial or full name of the chart/procedure.'\}\}\\
4. Name: charts\_afd\_get\\
Description: Get the AFD for a specified airport\\
Parameters: \{'apt': \{'type': 'string', 'description': 'FAA or ICAO airport identifier (KCLT or CLT). Required is true.', 'required': True\}\}\\
5. Name: preferred-routes\_get\\
Description: Get all of the preferred routes\\
Parameters: \{\}\\
\textbf{</tools>}\\
\\
\textbf{Input}\\
User: I'm planning a trip to Chicago next week. Can you check the weather conditions at O'Hare International Airport (ICAO: ORD) for me? Also, I'd like to know the runway length and the type of surface of the longest runway there.\\
\\
\textbf{Your Response}: \textbf{<tool\_call>}[ \{"name": "weather\_metar\_get", "parameters": \{"apt": "ORD"\} \}, \{"name": "airports\_get", "parameters": \{"apt": "ORD"\} \} ] \textbf{</tool\_call>}
\end{tcolorbox}
\caption{Sample example from ToolAlpaca test data.}
\label{tab:toolalpaca}
\end{figure*}

\end{document}